\documentclass[11pt]{article}

\usepackage[final]{acl}

\usepackage{times}
\usepackage{latexsym}

\usepackage[T1]{fontenc}

\usepackage[utf8]{inputenc}

\usepackage{microtype}

\usepackage{inconsolata}

\usepackage{graphicx}

\usepackage{subfigure}
\usepackage{caption}
\usepackage{float}
\usepackage{bm}
\usepackage{diagbox}
\usepackage{algorithm}
\usepackage{algorithmicx}
\usepackage{algpseudocode}
\usepackage{url}
\usepackage{indentfirst}
\usepackage{verbatim}
\usepackage{booktabs,multirow}
\usepackage{tikz}
\usetikzlibrary{spy}
\usepackage{cuted}
\usepackage{makecell}
\usepackage{bbding}
\usepackage{pifont}
\usepackage{array}
\usepackage{amsmath}

\hypersetup{hidelinks}
\usepackage{amssymb}
\algnewcommand\algorithmicinput{\textbf{Input:}}
\algnewcommand\Input{\item[\algorithmicinput]}
\algnewcommand\algorithmicoutput{\textbf{Output:}}
\algnewcommand\Output{\item[\algorithmicoutput]}
\algnewcommand{\Continue}{\State \textbf{continue}}
\usepackage[table]{xcolor}
\definecolor{mygray}{gray}{.92}
\definecolor{mypink}{RGB}{255, 182, 193}
\definecolor{darkgreen}{rgb}{0.0, 0.5, 0.0} 

\usepackage[table]{xcolor}
\usepackage{color}

\title{Noisy Test-Time Reinforcement Learning for Code LLMs}

\author{
 \textbf{Xikai Yang\textsuperscript{1}},
 \textbf{Hieu Trung Nguyen\textsuperscript{1}},
 \textbf{Dunyuan Xu\textsuperscript{1}},
 \textbf{Yuzhi Zhao\textsuperscript{2}},
\\
 \textbf{Jinpeng Li\textsuperscript{1}},
 \textbf{Wenao Ma\textsuperscript{1}}\thanks{Corresponding author.},
 \textbf{Pheng-Ann Heng\textsuperscript{1}}
\\
\\
 \textsuperscript{1}The Chinese University of Hong Kong,
 \textsuperscript{2}Huazhong University of Science and Technology \\
 \texttt{\href{mailto:yangxk@link.cuhk.edu.hk}{yangxk@link.cuhk.edu.hk}}
}

\begin{document}
\maketitle
\begin{abstract}
Large language models (LLMs) have demonstrated remarkable performance across various code-related tasks. However, unlike carefully curated datasets that are typically high-quality and error-free, real-world user instructions are often vague and error-prone, posing significant challenges to the robustness of code LLMs.
Furthermore, robustness-oriented fine-tuning relies on paired clean-noisy samples, which are costly to curate and require sophisticated noisy simulation techniques.
To address these challenges, we propose the Noisy Test-time Reinforcement Learning framework (NTRL-Code), which enables robust self-evolution of code LLMs using only unlabeled noisy data during the testing stage.
Specifically, NTRL-Code uses conservative self-denoising to obtain a cleaner semantic anchor for target estimation, and employs an abstract-syntax-tree (AST)-based structural aggregation mechanism to estimate a proxy target from multiple candidate programs. The policy is then optimized on the original noisy prompts with a hybrid reward that combines format validity, code similarity, and anti-repetition signals.
Extensive experiments on three benchmarks, each incorporating character-level, word-level, and paragraph-level perturbations,
demonstrate that NTRL-Code yields robust and consistent improvements, stabilizing the predictions of various base models. Our code is available at \href{https://github.com/Xikai97/NTRL-Code}{https://github.com/Xikai97/NTRL-Code}.
\end{abstract}

\section{Introduction}
\begin{figure}[t]
  \includegraphics[width=0.48\textwidth]{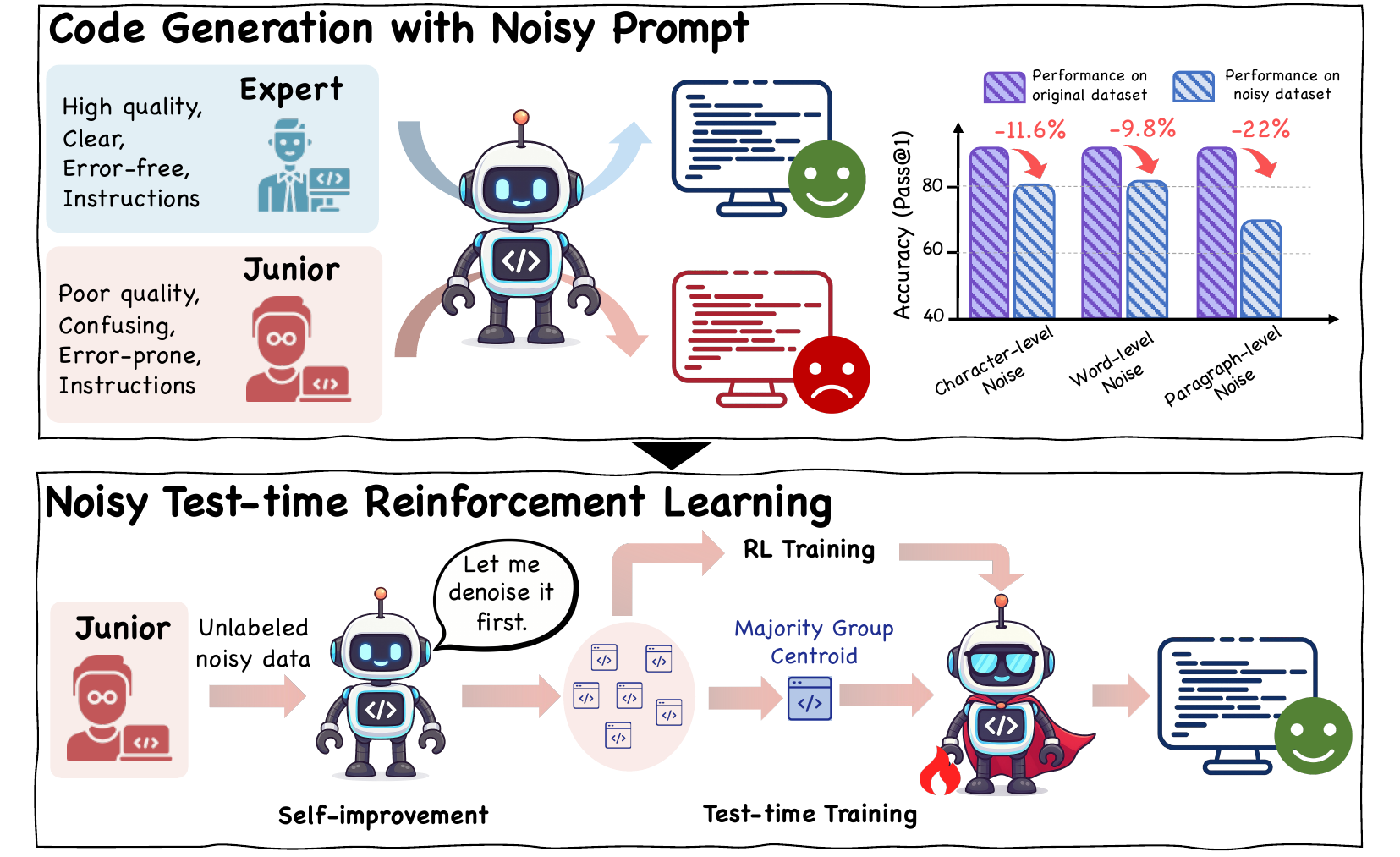} 
  \caption{Impact of noisy prompts on Qwen2.5-coder-32B evaluated on HumanEval (top). To address the challenge of noisy inputs, we propose a noisy test-time reinforcement learning framework that trains exclusively on unlabeled noisy data (bottom).}
  \label{fig:teaser}
\end{figure}

\noindent Recently, LLMs have become a prominent research focus, driving advancements across various domains, including software engineering~\cite{jin2024llms}, 
robotics~\cite{kim2024survey}, 
healthcare~\cite{goyal2024healai,shool2025systematic}. Among these applications, code generation stands out as one of the most significant tasks, where LLMs demonstrate the potential to achieve, or even surpass, human-level intelligence. Several widely used LLMs, including Claude series models~\cite{anthropic_claude_sonnet_4_5}, GPT series models~\cite{openai2025gpt5}, and Qwen series models~\cite{qwen3technicalreport} have proven exceptional performance in generating code snippets given specific textual instructions.
However, previous studies have shown that the quality of the generated code of LLMs is highly dependent on the precision of the input instruction prompt~\cite{chen2026nlperturbator,paleyes2026code,zhuo2025bigcodebench}. Even minor changes, such as altering a few words or rephrasing the prompt with equivalent descriptions, can result in a significant drop in performance (As shown in the top of Figure~\ref{fig:teaser}). 
Moreover, real-world users often have varying levels of expertise. For instance, compared with expert, junior users may provide vague or erroneous instructions, posing new challenges to the performance of code LLMs.

Despite the existence of various approaches aimed at enhancing the robustness of LLMs~\cite{sun2024evaluating,zhao2024improving,agrawal2025enhancing}, a large body of prior robustness works rely on supervised training with high-quality noisy-clean data pairs, which imposes a substantial burden on dataset preparation~\cite{zhao2024improving,bai2025enhancing} and often leads to overfitting on specific noise patterns.
Another category of approaches focuses on aligning features between clean and noisy samples and calibrating hidden layer embeddings~\cite{sun2024evaluating, liu2025improving, liu2025reducing}. Yet, directly modifying the parameters of specific hidden layers risks disrupting the reasoning process of LLMs, potentially leading to degraded performance.
Other works propose to incorporate self-denoising~\cite{agrawal2025enhancing} in the framework and leverage the model's inherent denoising capabilities through prompt engineering to iteratively refine noisy inputs.
The main concern with this method is that it fails to fundamentally improve the model's robustness. Moreover, the iterative denoising process significantly increases inference time.
Therefore, a label-efficient approach to enhance the performance of Code LLMs against various types of perturbations is highly desired.

Two major classes of methods focus on such model evolution without supervision: (1) entropy-based methods, such as Tent~\cite{wang2020tent}, EMPO~\cite{zhang2026right}, and RENT~\cite{prabhudesai2025maximizing}; (2) self-rewarding methods, like Test-Time Reinforcement Learning (TTRL)\cite{zuo2026ttrl}, TTRV\cite{singh2026ttrv}, etc.
However, both types of methods lack specific designs to handle perturbations within the prompt. Noisy inputs can negatively affect entropy variation and degrade the quality of self-reward mechanisms. 
Moreover, compared with the mathematical datasets typically used in these methods, coding problems present additional complexity in identifying an appropriate reward for self-supervision, particularly for approaches relying on majority voting. For instance, two code snippets with different variable names or minor formatting differences but identical execution logic can still be grouped together. This makes identifying the most frequent output, as commonly applied in mathematical problems~\cite{zuo2026ttrl}, unsuitable for coding tasks.

Therefore, in this work, we propose \textbf{NTRL-Code}, a novel test-time training framework designed to steadily improve code LLMs' performance using only unlabeled, noise-perturbed test samples.
Specifically, NTRL-Code incorporates a self-improvement module that performs a self-denoising call before each rollout to mitigate the impact of perturbed prompts.
Subsequently, a dual-rollout is conducted on both the original noisy prompt and denoised prompt to decouple reward calculation from proxy-target estimation.
Additionally, NTRL-Code leverages the Abstract Syntax Tree (AST) to parse each code snippet and employs a majority voting mechanism to generate pseudo-labels.
Finally, we design a hybrid reward system that combines format reward, code similarity reward, and no-repetition reward, enabling the policy model to be updated in a code-friendly manner.
Our contributions are summarized as follows: \\
1. We formulate noisy test-time reinforcement learning for code generation, where a code LLM adapts from unlabeled noisy prompts without clean-noisy pairs or executable feedback. \\
2. We propose \textbf{NTRL-Code}, a decoupled denoising-optimization framework that estimates proxy targets from a denoised branch while optimizing the policy on the original noisy branch. \\
3. We introduce structure-aware proxy target estimation based on AST canonicalization and clustering, together with a hybrid reward for stable test-time RL. \\
4. Experiments on three benchmarks under diverse perturbations show consistent gains across multiple base models, validating the effectiveness of our design.

\section{Related Work}

\subsection{LLM Robustness}
\noindent With the remarkable performance achieved by LLMs across various fields, recent studies have begun to explore the impact of input perturbations on generated outputs~\cite{nookala2023adversarial,rauba2024quantifying,chaudhary2024towards,alahmari2025large,wang2025truth}. Given that diverse noise is likely to arise in real-world deployments, it is crucial to evaluate the robustness of LLMs and develop reliable AI systems to ensure resilience against such perturbations.
\citet{zhu2024promptbench} systematically built a library for LLM evaluation, which also supports research on adversarial prompt attacks at various levels, such as character-level and semantic-level attacks.  
\citet{zhu2023promptrobust} employed adversarial textual attacks to generate noisy prompts for LLM evaluation and demonstrated that LLMs are not robust to such adversarial prompts.  
\citet{chatterjee2024posix} developed a Prompt Sensitivity Index to measure the sensitivity of LLMs to variations in prompts.  
To enhance the robustness of LLMs, \citet{agrawal2025enhancing} compared several prevailing approaches and found that iterative self-denoising is effective in improving generation quality when dealing with noisy prompts. Other approaches, such as hidden state alignment~\cite{zhao2024improving}, adversarial training~\cite{bukharin2025adversarial}, and prompt optimization~\cite{wu2026evorefuse}, have also been explored in recent works to mitigate the impact of perturbations in instructions.

\subsection{AI coding Agent}
\noindent The advancement of AI coding agents has brought a revolution to programming users. Nearly all successful large models claim superior performance in solving coding problems~\cite{guo2024deepseek,hui2024qwen2,qwen3technicalreport,openai2025gpt5}.  
Since coding tasks are highly user-interactive and impose rigorous demands on the output, evaluating the robustness of coding agents has gradually emerged as a critical research direction~\cite{yang2024robustness,liu2025adversarial,lin2025robunfr,chen2025dynamic}.  
However, few studies have focused on enhancing the robustness of coding agents.  
\citet{hossen2024adversarial} leveraged GPT models to generate adversarial prompts and constructed corresponding instruction-tuned datasets for coding agent training.  
\citet{liu2025improving} proposed a model editing approach that updated the most unstable layers during supervised training on paired original and perturbed prompts.

\begin{figure*}[t]
  \includegraphics[width=0.97\textwidth]{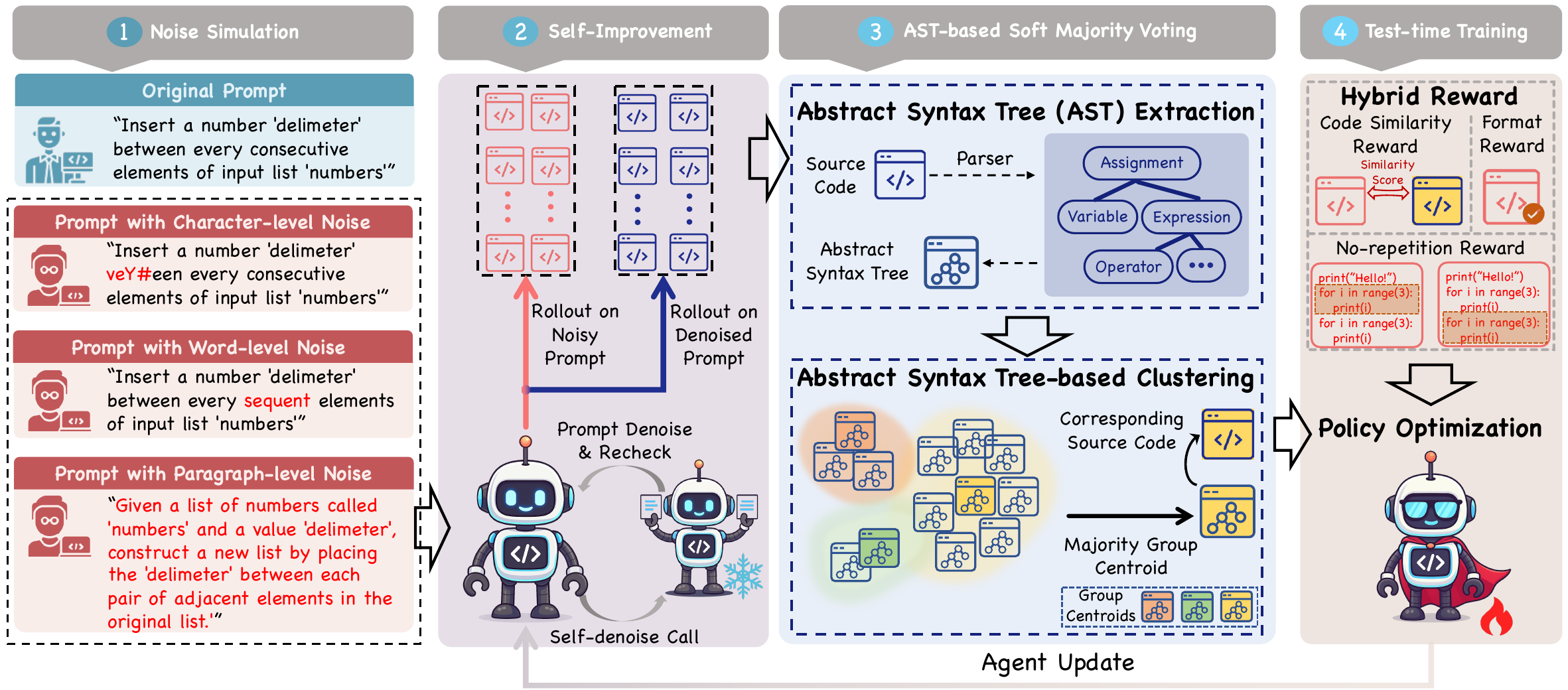} 
  \caption{(1) Examples of three types of simulated perturbation and (2-4) an overview of the proposed NTRL-Code framework. The framework consists of three key modules: (2) a self-improvement module for initial prompt correction to improve proxy-target estimation; (3) an Abstract Syntax Tree–based majority voting mechanism to identify the centroid of the largest group for reward calculation; and (4) policy optimization using a hybrid reward signal to enhance the model with unlabeled data.}
  \label{fig:main_framework}
\end{figure*}

\section{Methodology}

\subsection{Overview of NTRL-Code Framework}
\noindent To address the performance degradation in code generation caused by noisy instructions, we propose the NTRL-Code framework to improve model's performance by using only unlabeled noisy data.
Figure~\ref{fig:main_framework} provides an overview of the NTRL-Code framework.
Specifically, we first utilize the inherent denoising capability of the current policy model to rectify input noisy prompt. 
Afterward, two rollouts are performed separately: one based on the original noisy prompt and the other on the denoised prompt. 
Next, we perform AST-based structural aggregation on the denoised rollouts and select the representative program from the largest structural cluster as a proxy target, which is a label-free estimate of the most stable program structure under the denoised semantic intent. Finally, we update the policy using GRPO on candidate programs generated from the original noisy prompt. The reward compares noisy-branch candidates against the proxy target through a hybrid reward function.

\subsection{Prompt Correction via Self-denoising}
\noindent Unlike previous work~\cite{agrawal2025enhancing}, which employed self-denoising to iteratively refine noisy prompts and improve inference results, NTRL-Code utilizes self-denoising solely to enhance the reliability of proxy target estimation from a group of candidates.
This design ensures that self-denoising in NTRL-Code does not interfere with the actual inference stage.
Furthermore, it directly contributes to model evolution during the test-time training phase, enhancing the model's inherent robustness.

Figure~\ref{fig:pt_SD_ntrl} in the Appendix illustrates the prompt used to call the current policy model $\pi_{\theta}(\cdot|x)$ to denoise input perturbed instructions. More concretely, during the denoising process, we temporarily modify the system prompt, transforming its role from a code generator to an instruction denoiser.
However, for certain models with limited instruction-following capabilities, such as DeepSeek-1.3B-Coder, it is challenging for them to accurately identify errors and effectively correct input prompts. More critically, they may generate excessive redundant content, leading to distribution shifts during the RL update process.
To mitigate such adverse effects, we apply a conservative recheck mechanism before using the denoised prompt for target estimation. Instead of relying on a semantic similarity model, we use a lightweight length-change ratio ($r_{\mathrm{change}}$) to detect overly aggressive rewrites.
\begin{equation}
\begin{aligned}
r_{\mathrm{change}} =
\frac{|\mathrm{len}(p)-\mathrm{len}(p_{\mathrm{denoise}})|}
{\max(\mathrm{len}(p),1)} .
\end{aligned}
\label{eq:denoise}
\end{equation}
We set the failure threshold to \(\tau_{\mathrm{len}}=0.5\). If \(r_{\mathrm{change}}>\tau_{\mathrm{len}}\), the denoising result is considered unreliable and we revert to the original noisy prompt as shown in Eq.~\eqref{eq:denoise_2}, where $D(\cdot, \phi)$ denotes the self-denoising operator with $\phi$ as the denoising prompt. $p$, $p_{\mathrm{denoise}}$, $\tilde{p}$ denote the original noisy prompt, the denoised prompt, and the final output prompt after recheck, respectively.
\begin{equation}
\begin{aligned}
p_{\mathrm{denoise}} = D(p,\phi), \;
\tilde{p} =
\begin{cases}
p, \text{if } r_{\mathrm{change}}>\tau_{\mathrm{len}},\\
p_{\mathrm{denoise}}, \text{otherwise}.
\end{cases}
\end{aligned}
\label{eq:denoise_2}
\end{equation}
This rule is designed to filter out destructive denoising cases, such as over-deletion of constraints, over-expansion with hallucinated requirements, or large semantic drift caused by excessive rewriting.
We then perform two separate rollouts: noisy rollouts conditioned on the original prompt \(p\), and target-estimation rollouts conditioned on the rechecked prompt \(\tilde{p}\). This decoupling ensures that policy optimization is performed on the noisy input distribution, while the proxy target is estimated from a cleaner and more stable prompt distribution.

\subsection{AST-based Structural Aggregation}
\label{sec:ast}
\noindent To construct a label-free reward signal for code generation, we estimate a proxy target from multiple programs sampled from the rechecked denoised prompt $\tilde{p}$. Since programs with different variable names, comments, or formatting may implement the same structure, raw textual majority voting is unsuitable for code. We therefore perform structure-aware aggregation based on Python ASTs.
For each candidate program $\tilde{y}_i$ generated from rechecked prompt $\tilde{p}$, we first extract the code block and attempt to parse it into a Python AST. If parsing succeeds, we canonicalize the tree by replacing user-defined variable names, function names, and argument names with deterministic placeholders. This normalization reduces spurious differences caused by renaming. We then serialize the canonicalized tree using ast.dump() function (excluding attributes and annotations) and compute an MD5 hash of the resulting string as its structural fingerprint. Candidates with the same structural fingerprint are assigned to the same cluster.
Some generated programs may be syntactically incomplete and therefore cannot be parsed into an AST. For these cases, we use a lexical fallback. Specifically, we normalize the code by removing comments, condensing whitespace, and lowercasing text, then compute token-level Jaccard similarity. Two unparsable candidates are assigned to the same cluster if their Jaccard similarity is at least \(\epsilon=0.92\). We selected \(\epsilon\) by a small parameter sweep over candidate thresholds and used the value that achieved the best Pass@1.
After clustering, we select the largest cluster and use its representative candidate as the proxy target $\tilde{y}^{*}$. 
Algorithm~\ref{alg:ast_aggregation} summarizes the procedure, where $\mathrm{ExtractCode}$, $\mathrm{AST}$, and $\mathrm{Canon}$ denote the extraction, parsing, and canonicalization steps described above.

\begin{algorithm}[!h] 
\caption{AST-based Structural Aggregation}
\label{alg:ast_aggregation}
\begin{algorithmic}[1]
    \Input Candidate set $\mathcal{Y}=\{\tilde{y}_1, \tilde{y}_2, \cdots, \tilde{y}_N\}$; lexical threshold $\epsilon$
    \Output Proxy target $\tilde{y}^{*}$

    \State $\mathcal{C} \leftarrow \varnothing$; $\mathcal{M} \leftarrow \varnothing$
    
    \For{$i \leftarrow 1 \textbf{ to } N$}
        \State $c_i \leftarrow \mathrm{ExtractCode}(\tilde{y}_i)$
        \If{$c_i$ is parsable}
          \State $h_i \leftarrow \mathrm{MD5}\big(\mathrm{Canon}(\mathrm{AST}(c_i))\big)$
          \State $k \leftarrow$ the cluster with key $h_k = h_i$
        \Else 
          \State $T_i \leftarrow \mathrm{Tok}(\mathrm{Nor}(c_i))$
          \State $k \leftarrow$ first lexical cluster with $\mathrm{Jac}(T_i, T_k) \ge \epsilon$
        \EndIf
        \If{$k$ exists}
          \State $C_k \leftarrow C_k \cup \{i\}$
        \Else 
          \State $C_{\text{new}} \leftarrow \{i\}$ 
          \State $\mathcal{C} \leftarrow \mathcal{C} \cup \{C_{\text{new}}\}$; 
        \EndIf
    \EndFor
    \State $k_{\max} \leftarrow \arg\max_k |C_k|$
    \State $i^{*} \leftarrow$ index of the first candidate assigned to the largest cluster $C_{k_{\max}}$
    \State $\tilde{y}^{*} \leftarrow \tilde{y}_{i^{*}}$
    \State \Return $\tilde{y}^{*}$
\end{algorithmic}
\end{algorithm}

\subsection{Test-time RL with Hybrid Rewards}
\noindent Similar to a standard RL training framework, test-time RL also updates the policy model based on reward signals.
These reward signals are computed by comparing proxy target $\tilde{y}^{*}$ with the candidate rollout responses $\{y_1, y_2, \cdots, y_N\}$ generated from the original noisy prompt by the current policy model.
In particular, we design a hybrid reward including three parts to reduce the risk of optimizing toward a single brittle signal.
\\
\textbf{\textit{Format reward.}} We determine whether a non-empty code block can be successfully extracted from the response; if successful, $R_{format}=1$, otherwise $R_{format}=0$. \\
\textbf{\textit{Code Similarity Reward.}} Let the extracted code for the candidate sample $y_i$ and the proxy target $\tilde{y}^{*}$ be denoted as $C_i$ and $C_{gt}$, respectively. The similarity score is computed as the sum of the normalized Levenshtein distance~\cite{lcvenshtcin1966binary} and the Jaccard similarity~\cite{kosub2019note}.  
Specifically, let $Nor(\cdot)$ represent the normalization process, which removes comments, condenses spaces, and converts text to lowercase. The normalized code can then be expressed as $C'_{i} = Nor(C_i)$ and $C'_{gt} = Nor(C_{gt})$. The normalized Levenshtein similarity reward is computed as shown in Eq.~\eqref{eq:lev_sim}.  
Similarly, let $Tok(\cdot)$ denote the tokenization operation. Using this, $T_i = \mathrm{Tok}(C_i)$ and $T_{gt} = \mathrm{Tok}(C_{gt})$ represent the token sets. The Jaccard similarity reward is then calculated as shown in Eq.~\eqref{eq:jac_sim}.  
The final fused similarity reward is computed as Eq.~\eqref{eq:sim_overall}.
\begin{equation}
\begin{aligned}
    r_{lev} = 1 - \min \big(1, \frac{Lev(C'_{i}, C'_{gt})}{\max(1, |C'_{i}|, |C'_{gt}|)}\big).
\end{aligned}
\label{eq:lev_sim}
\end{equation}
\begin{equation}
\begin{aligned}
    r_{jac} \;=\; \dfrac{|T_i \cap T_{gt}|}{\max\!\{1,\; |T_i \cup T_{gt}|\}}.
\end{aligned}
\label{eq:jac_sim}
\end{equation}
\begin{equation}
\begin{aligned}
    R_{sim} = 0.5 \cdot r_{lev} + 0.5 \cdot r_{jac}.
\end{aligned}
\label{eq:sim_overall}
\end{equation}
\textbf{\textit{No-repetition Reward.}} We utilize sliding windows to measure the extent of repetition within the generated code snippet. Let $P' = Nor(P)$ represent the processed input prompt. The window length $W$ and step size $S$ are defined as shown in Eq.~\eqref{eq:slide_w_s}.
\begin{equation}
\begin{aligned}
    W = \max\!\Bigl(&10,\; \min\!\bigl(80,\; \bigl\lfloor |P'|/10 \bigr\rfloor + 10 \bigr)\Bigr), \\
S &= \max\!\Bigl(5,\; \bigl\lfloor W/2 \bigr\rfloor\Bigr).
\end{aligned}
\label{eq:slide_w_s}
\end{equation}
We then define the number of overlaps $M$ by applying sliding windows of size $W$ over $P'$ with stride $S$. The normalized overlap ratio $M'$ is computed as shown in Eq.~\eqref{eq:slide_r}. The reward for no-repetition is defined as: $R_{rep} = 1 - M'.$
\begin{equation}
\begin{aligned}
M' \;=\; \frac{M}{\,1 + \left\lfloor |P'|/S \right\rfloor\, }.
\end{aligned} 
\label{eq:slide_r}
\end{equation}
\textbf{\textit{Hybrid Reward.}}  
The final hybrid reward is given as in Eq.~\eqref{eq:total_reward}, where we let $(\alpha, \beta)=(0.5,0.25)$.
Specifically, format reward filters structurally invalid responses, similarity reward aligns noisy-branch candidates with the proxy target, and no-repetition reward penalizes degenerate repetitions that frequently appear under noisy prompts. The complete update is driven by both code validity and structural consistency rather than raw textual matching alone.
\begin{equation}
\begin{aligned}
R_{total} = \alpha R_{sim} + \beta R_{format} + (1-\alpha-\beta)R_{rep}.
\end{aligned} 
\label{eq:total_reward}
\end{equation}

\subsection{Why Decoupled Test-Time RL Improves Robustness} \label{sec:why_ttrl}
\noindent NTRL-Code differs from generic test-time self-improvement in the source of its reward anchor. In standard TTRL, both the candidate rollouts and the pseudo-target are obtained from the same input distribution, which makes the reward signal itself noise-dependent under noisy prompts. If the perturbation changes the model's interpretation, majority voting may reinforce a perturbed or spurious pattern.
NTRL-Code instead optimizes the policy under a decoupled objective. Let $p$ and $\tilde{p}$ denote the original noisy prompt and the rechecked denoised prompt, respectively. The proxy target is estimated from denoised rollouts,
\begin{equation}
\tilde{y}^{*} = A(\{\tilde{y}_i\}_{i=1}^{N}), \quad \tilde{y}_i \sim \pi_{\theta}(\cdot|\tilde{p}),
\label{eq:explan_2}
\end{equation}
where \(A(\cdot)\) denotes AST-based structural aggregation. The policy update is then performed on noisy-branch candidates ($y_i \sim \pi_{\theta}(\cdot|p)$) using rewards $R(y_i, \tilde{y}^{*})$. This yields the adaptation objective as shown in Eq.~\eqref{eq:explan_1}.
\begin{equation}
\max_{\theta}
\mathbb{E}_{p \sim \mathcal{D}}
\mathbb{E}_{y \sim \pi_{\theta}(\cdot|p)}
\left[
R(y, A(\pi_{\theta}(\cdot|\tilde{p})))
\right].  
\label{eq:explan_1}
\end{equation}
This objective encourages outputs conditioned on noisy prompts to match a semantic anchor estimated from a cleaner prompt distribution. Therefore, the update pushes the model toward noise-invariant mappings rather than simply increasing confidence on the noisy input itself.

This design also explains why NTRL-Code is less likely to overfit to a specific perturbation pattern. The optimization target is not derived from the noisy text alone. It is anchored by the denoised branch and aggregated in a structure-aware code space. Moreover, the small number of update steps constrains the magnitude of adaptation. The cross-perturbation and cross-dataset results in Section~\ref{sec:general_analysis} further suggest that the learned update generalizes beyond the perturbation type used during adaptation.

\begin{table*}[!h]
\centering
\setlength{\tabcolsep}{2.3pt}
\renewcommand{\arraystretch}{1.15} 
\scalebox{0.76}{
\begin{tabular}{>{\centering\arraybackslash}m{2.0cm}|>{\centering\arraybackslash}m{2.3cm}|ccc|ccc|ccc}
\toprule
\multirow{2}{=}{\centering\textbf{Base Model}} & \multirow{2}{=}{\centering\textbf{Methods}} & \multicolumn{3}{c|}{\textbf{HumanEval}}  & \multicolumn{3}{c|}{\textbf{LeetCode Bench}}  & \multicolumn{3}{c}{\textbf{MBPP}}\\
\cmidrule(lr){3-11}
& & \textit{Char-Perb}    & \textit{Word-Perb}   & \textit{Para-Perb}     & \textit{Char-Perb}    & \textit{Word-Perb}   & \textit{Para-Perb}  & \textit{Char-Perb}    & \textit{Word-Perb}   & \textit{Para-Perb}  \\
\midrule
\multirow{5}{=}{\centering{DeepSeek-Coder-1.3B}} &Original &$\text{53.0}$ &$\text{54.3}$ &$\text{56.1}$  &$\text{3.3}$ &$\text{5.0}$ &$\text{5.0}$ &$\text{44.6}$ &$\text{43.8}$ &$\text{46.6}$  \\
&Few-shot &$\text{49.4}_{-3.6}$ &$\text{56.1}_{+1.8}$ &$\text{56.1}_{+0.0}$ &$\text{3.3}_{+0.0}$ &$\textbf{6.7}_{+1.7}$ &$\text{5.0}_{+0.0}$  &$\text{48.2}_{+3.6}$ &$\text{46.0}_{+2.2}$ &$\text{47.8}_{+1.2}$ \\
&Big-Denoise &$\text{54.3}_{+1.3}$ &$\text{50.0}_{-4.3}$ &$\text{37.2}_{-18.9}$ &$\text{7.8}_{+4.5}$ &$\text{5.6}_{+0.6}$ &$\text{4.4}_{-0.6}$  &$\text{48.4}_{+3.8}$ &$\text{44.0}_{+0.2}$ &$\text{0.2}_{-46.4}$ \\
&Iter-SD &$\text{51.8}_{-1.2}$ &$\text{52.4}_{-1.9}$ &$\text{56.1}_{+0.0}$ &$\text{3.3}_{+0.0}$ &$\text{6.1}_{+1.1}$ &$\text{5.0}_{+0.0}$  &$\text{48.6}_{+4.0}$ &$\text{47.0}_{+3.2}$ &$\text{48.6}_{+2.0}$ \\
\rowcolor{mygray}
\cellcolor{white}&NTRL-Code &$\textbf{55.5}_{+2.5}$ &$\textbf{59.1}_{+4.8}$ &$\textbf{62.2}_{+6.1}$ &$\textbf{7.2}_{+3.9}$ &$\textbf{6.7}_{+1.7}$ &$\textbf{8.9}_{+3.9}$  &$\textbf{49.2}_{+4.6}$ &$\textbf{48.6}_{+4.8}$ &$\textbf{50.6}_{+4.0}$ \\
\midrule
\multirow{5}{=}{\centering{DeepSeek-Coder-6.7B}} &Original  &$\text{62.8}$ &$\text{64.0}$ &$\text{66.5}$ &$\text{16.1}$ &$\text{15.0}$ &$\text{16.1}$ &$\text{60.0}$ &$\text{57.8}$ &$\text{61.6}$  \\
&Few-shot &$\text{58.5}_{-4.3}$ &$\text{64.0}_{+0.0}$ &$\text{59.1}_{-7.4}$ &$\text{17.8}_{+1.7}$ &$\text{15.0}_{+0.0}$ &$\text{18.9}_{+2.8}$ &$\text{61.4}_{+1.4}$ &$\text{58.4}_{+0.6}$ &$\text{62.8}_{+1.2}$ \\
&Big-Denoise &$\text{67.1}_{+4.3}$ &$\text{63.4}_{-0.6}$ &$\text{49.4}_{-17.1}$ &$\text{16.1}_{+0.0}$ &$\textbf{18.3}_{+3.3}$ &$\text{16.1}_{+0.0}$ &$\text{60.4}_{+0.4}$ &$\text{58.3}_{+0.5}$ &$\text{54.0}_{-7.6}$ \\
&Iter-SD &$\text{67.1}_{+4.3}$ &$\text{62.8}_{-1.2}$ &$\text{65.2}_{-1.3}$ &$\text{17.2}_{+1.1}$ &$\text{17.2}_{+2.2}$ &$\text{17.8}_{+1.7}$ &$\textbf{62.8}_{+2.8}$ &$\text{51.6}_{-6.2}$ &$\text{62.0}_{+0.4}$ \\
\rowcolor{mygray}
\cellcolor{white}&NTRL-Code  &$\textbf{68.3}_{+5.5}$ &$\textbf{67.1}_{+3.1}$ &$\textbf{72.0}_{+5.5}$ &$\textbf{18.3}_{+2.2}$ &$\textbf{18.3}_{+3.3}$ &$\textbf{20.6}_{+4.5}$ &$\text{62.6}_{+2.6}$  &$\textbf{60.6}_{+2.8}$  &$\textbf{63.8}_{+2.2}$  \\
\midrule
\multirow{5}{=}{\centering{Qwen2.5-Coder-32B}} &Original  &$\text{79.9}$ &$\text{81.7}$ &$\text{69.5}$  &$\text{63.9}$ &$\text{63.3}$ &$\text{67.2}$ &$\text{80.0}$ &$\text{79.2}$ &$\text{78.2}$ \\
&Few-shot &$\text{80.5}_{+0.6}$ &$\text{81.1}_{-0.6}$ &$\textbf{73.2}_{+3.7}$ &$\text{62.2}_{-1.7}$ &$\text{66.7}_{+3.4}$ &$\text{66.7}_{-0.5}$ &$\text{19.2}_{-60.8}$ &$\text{64.8}_{-14.4}$ &$\text{50.6}_{-27.6}$ \\
&Big-Denoise &$\text{74.4}_{-5.5}$ &$\text{70.7}_{-11.0}$ &$\text{56.1}_{-13.4}$ &$\text{67.2}_{+3.3}$ &$\text{68.3}_{+5.0}$ &$\text{66.1}_{-1.1}$ &$\text{75.2}_{-4.8}$ &$\text{73.6}_{-5.6}$ &$\text{69.6}_{-8.6}$ \\
&Iter-SD &$\text{78.0}_{-1.9}$ &$\text{79.3}_{-2.4}$ &$\text{65.9}_{-3.6}$ &$\text{64.4}_{+0.5}$ &$\text{66.7}_{+3.4}$ &$\text{60.6}_{-6.6}$ &$\text{80.2}_{+0.2}$ &$\text{79.6}_{+0.4}$ &$\text{78.4}_{+0.2}$ \\
\rowcolor{mygray}
\cellcolor{white}&NTRL-Code  &$\textbf{81.7}_{+1.8}$ &$\textbf{82.9}_{+1.2}$ &$\text{72.0}_{+2.5}$ &$\textbf{71.1}_{+7.2}$ &$\textbf{71.7}_{+8.4}$ &$\textbf{68.9}_{+1.7}$ &$\textbf{82.6}_{+2.6}$ &$\textbf{80.6}_{+1.4}$ &$\textbf{78.8}_{+0.6}$ \\
\bottomrule
\end{tabular}
}
\caption{Method comparison of our proposed NTRL-Code and three additional label-free denoising approaches across three base models (DeepSeek-Coder-1.3B, 6.7B, and Qwen2.5-coder-32B). Few-shot, Big-Denoise, and Iter-SD denotes the Few-shot denoising, denoising using a large model, and iterative self-denoising, respectively. Pass@1 accuracy is reported for three types of perturbations across three benchmarks.}
\label{table:method_compare_all}
\end{table*}

\section{Experiments}
\noindent We conduct extensive experiments on three datasets and their perturbed variants (described in Appendix~\ref{sec:prelimiary}) to evaluate the performance of our framework. In the following sections, we first present a detailed explanation of the general experimental setup used in our study, followed by the main results of our method in comparison to three label-free approaches. Finally, we demonstrate the effectiveness of our framework's design through a series of ablation studies and evaluate the performance of NTRL-Code on the noisy dataset with mixed-type perturbations, as well as in cross-dataset and cross-perturbation settings.

\subsection{Experiment Setup}
\noindent We leverage GRPO~\cite{shao2024deepseekmath} to estimate the advantage function in RL. The training process uses a batch size of 8 and a learning rate of $5 \times 10^{-7}$. We conduct RL update over 100 steps, with 64 rollouts sampled per prompt to estimate proxy target and 32 rollouts for advantage computation.
To evaluate the performance of the proposed NTRL-Code framework, we use the standard Pass@1 metric, which measures the proportion of test samples successfully solved on the first attempt, closely reflecting typical user behavior~\cite{liu2025improving}.
We impose three levels of perturbation to prompts: character-level, word-level, and paragraph-level.
These perturbations cover common surface-form and paraphrastic variations.
Figure~\ref{fig:main_framework}-1 illustrates an example of our simulated perturbations. 
Details of the implementation of perturbation simulation along with additional examples can be found in the Appendix~\ref{sec:perb_details}.
We apply three types of noise to three datasets (LeetCode-Bench~\cite{guo2024deepseek}, HumanEval~\cite{chen2021evaluating}, and MBPP~\cite{austin2021program}), resulting in nine noisy benchmarks to evaluate the robustness of LLMs in code generation tasks. 

\subsection{Main Results}
\noindent We evaluate the NTRL-Code approach on three base models: DeepSeek-Coder-1.3B, DeepSeek-Coder-6.7B, and Qwen2.5-Coder-32B. For each base model, we compare the performance of NTRL-Code against three label-free denoising methods: Few-shot denoising~\cite{zhang2023certified}, denoising using a large model~\cite{qwen3technicalreport}, and iterative self-denoising~\cite{agrawal2025enhancing}.  
Specifically, few-shot denoising leverages a small set of clean-noisy examples to effectively guide the model denoise input prompt and then do inference on the denoised prompt. 
The specific prompt for few-shot denoising is given in the Appendix~\ref{sec:prompt_summary}. 
For denoising using a large model, we employ \textit{Qwen3-Coder-480B-A35B-Instruct} as an external powerful denoiser with a carefully designed prompt (shown in Appendix~\ref{sec:prompt_summary}) to modify noisy prompt.  
For iterative self-denoising, we adopt the test-time scaling strategy, where self-denoising is performed iteratively until the content remains unchanged or the maximum number of iterations is reached.

Table~\ref{table:method_compare_all} presents the corresponding main results. 
Our proposed NTRL-Code framework consistently improves performance across all three datasets and all perturbation levels relative to the original models. Averaged over the nine settings per model, the performance gains are $4.0\%$, $3.5\%$, and $3.0\%$ for DeepSeek-Coder-1.3B, DeepSeek-Coder-6.7B, and Qwen2.5-Coder-32B, respectively. Averaged by perturbation type, NTRL-Code yields gains of $3.7\%$, $3.5\%$, $3.4\%$ under character, word, and paragraph-level perturbations, respectively.
In contrast, Few-shot denoising exhibits highly unstable behavior across datasets and perturbation types. For example, it improves HumanEval under paragraph-level perturbation by 3.7\% with the Qwen2.5-Coder-32B base model but causes severe degradation on MBPP dataset. The Big-Denoise approach delivers improvements in some noisy settings but is particularly vulnerable to paragraph-level perturbations, resulting in an average drop of 12.6\%. Iterative self-denoising is relatively more robust than the other baselines, yet its overall performance remains inferior to NTRL-Code. 
These results reveals that NTRL-Code exhibits strong robustness in enhancing model performance against perturbations using only unlabeled noisy samples and a few RL updates.

\begin{table}[!h]
\centering
\setlength{\tabcolsep}{2.3pt}
\renewcommand{\arraystretch}{1.15} 
\scalebox{0.83}{
\begin{tabular}{>{\centering\arraybackslash}m{3.5cm}|cccccc|ccc}
\toprule
\multirow{2}{=}{\centering\textbf{Methods}} & \multicolumn{3}{c}{\textbf{HumanEval}} \\
\cmidrule(lr){2-4}
& \textit{Char-Perb}    & \textit{Word-Perb}   & \textit{Para-Perb}   \\
\midrule
Original &$\text{53.0}$ &$\text{54.3}$ &$\text{56.1}$  \\
NTRL-Code w/o SD &$\text{51.2}_{-1.8}$ &$\text{55.5}_{+1.2}$ &$\text{59.1}_{+3.0}$ \\
NTRL-Code w/ SD &$\text{55.5}_{+2.5}$ &$\text{59.1}_{+4.8}$ &$\text{62.2}_{+6.1}$ \\
\midrule
\multirow{2}{=}{\centering\textbf{Methods}}  & \multicolumn{3}{c}{\textbf{LeetCode Bench}} \\
\cmidrule(lr){2-4}
& \textit{Char-Perb}    & \textit{Word-Perb}   & \textit{Para-Perb}   \\
\midrule
Original &$\text{3.3}$ &$\text{5.0}$ &$\text{5.0}$ \\
NTRL-Code w/o SD &$\text{5.6}_{+2.3}$ &$\text{4.4}_{-0.6}$ &$\text{8.3}_{+3.3}$ \\
NTRL-Code w/ SD &$\text{7.2}_{+3.9}$ &$\text{6.7}_{+1.7}$ &$\text{8.9}_{+3.9}$ \\
\midrule
\multirow{2}{=}{\centering\textbf{Methods}} & \multicolumn{3}{c}{\textbf{MBPP}}\\
\cmidrule(lr){2-4}
& \textit{Char-Perb}    & \textit{Word-Perb}   & \textit{Para-Perb}  \\
\midrule
Original &$\text{44.6}$ &$\text{43.8}$ &$\text{46.6}$  \\
NTRL-Code w/o SD &$\text{47.6}_{+3.0}$ &$\text{45.2}_{+1.4}$ &$\text{47.8}_{+1.2}$ \\
NTRL-Code w/ SD  &$\text{49.2}_{+4.6}$ &$\text{48.6}_{+4.8}$ &$\text{50.6}_{+4.0}$ \\
\bottomrule
\end{tabular}
}
\caption{Self-denoising ablation study for the DeepSeek-Coder-1.3B model. SD refers to self-denoising.}
\label{table:SD_ablation}
\end{table}

\vspace{-0.1in}
\subsection{Self-Denoising Ablation}
\noindent We further investigate the effectiveness of the self-denoising component by removing the self-denoising call and the rollout on the denoised prompt. We evaluate NTRL-Code on the DeepSeek-Coder-1.3B model across all three datasets, and the results are summarized in Table~\ref{table:SD_ablation}. After ablating the self-denoising component, performance consistently drops (in some cases even falls below the original model). 
The average performance reductions at the character, word, and paragraph-level perturbations are $2.5\%$, $3.1\%$, and $2.1\%$, respectively. 
These findings suggest that self-denoising is particularly effective at correcting lower-level errors (e.g., spelling and tokenization issues) and provides stable performance gains. 
However, addressing content-level or semantic variations may require additional training to further enhance reasoning capabilities.
This ablation supports the decoupling hypothesis in Section~\ref{sec:why_ttrl}. When the denoised branch is removed, the proxy target is estimated under the same noisy distribution used for optimization, making the reward signal less stable.

\subsection{Rewards Ablation}
\noindent Here, we perform a detailed ablation study on the hybrid reward. Table~\ref{table:reward_ablation} reports results for seven distinct reward configurations on HumanEval with the DeepSeek-Coder-1.3B base model. For each configuration, we also report the performance change relative to the full hybrid reward.
Relying on a single reward component (\ding{172}, \ding{173}, \ding{174}) leads to clear degradations under three perturbation types. With two reward component variants (\ding{175}, \ding{176}, \ding{177}), we observe that paragraph-level perturbations are particularly sensitive to the missing of similarity reward and yields a $5.5\%$ drop, indicating that similarity reward is crucial to handle semantic variations. Overall, the best performance is achieved when all three reward components (Format, No-repetition, Code similarity) are enabled, underscoring the importance of the hybrid reward design.

\begin{table}[!h]
\centering
\setlength{\tabcolsep}{2.3pt}
\renewcommand{\arraystretch}{1.15} 
\scalebox{0.86}{
\begin{tabular}{c|c|c|c|ccc}
\toprule
& Format & Rep & Sim  & \textit{Char-Perb}    & \textit{Word-Perb}   & \textit{Para-Perb} \\
\midrule
\ding{172}&\ding{51}  &  &&$\text{52.4}_{-3.1}$ &$\text{53.7}_{-5.4}$ &$\text{57.9}_{-4.3}$ \\
\ding{173}& &\ding{51} & &$\text{51.8}_{-3.7}$ &$\text{55.5}_{-3.6}$ &$\text{57.3}_{-4.9}$ \\
\ding{174}& &  &\ding{51} &$\text{50.6}_{-4.9}$ &$\text{55.5}_{-3.6}$ &$\text{59.1}_{-3.1}$ \\
\ding{175}&\ding{51} &\ding{51}  & &$\text{53.0}_{-2.5}$ &$\text{56.7}_{-2.4}$ &$\text{56.7}_{-5.5}$ \\
\ding{176}&\ding{51} &  &\ding{51} &$\text{51.8}_{-3.7}$ &$\text{56.1}_{-3.0}$ &$\text{59.1}_{-3.1}$ \\
\ding{177}& &\ding{51} &\ding{51} &$\text{51.8}_{-3.7}$ &$\text{56.1}_{-3.0}$ &$\text{59.7}_{-2.5}$ \\
\midrule
\ding{178}&\ding{51} &\ding{51}  &\ding{51} &55.5 &59.1 &62.2 \\
\bottomrule
\end{tabular}
}
\caption{Reward ablation study on the HumanEval dataset with the base model DeepSeek-Coder-1.3B.}
\label{table:reward_ablation}
\vspace{-0.1in}
\end{table}

\subsection{Performance Evolution on RL Steps}
\begin{figure}[!h]
  \includegraphics[width=0.48\textwidth]{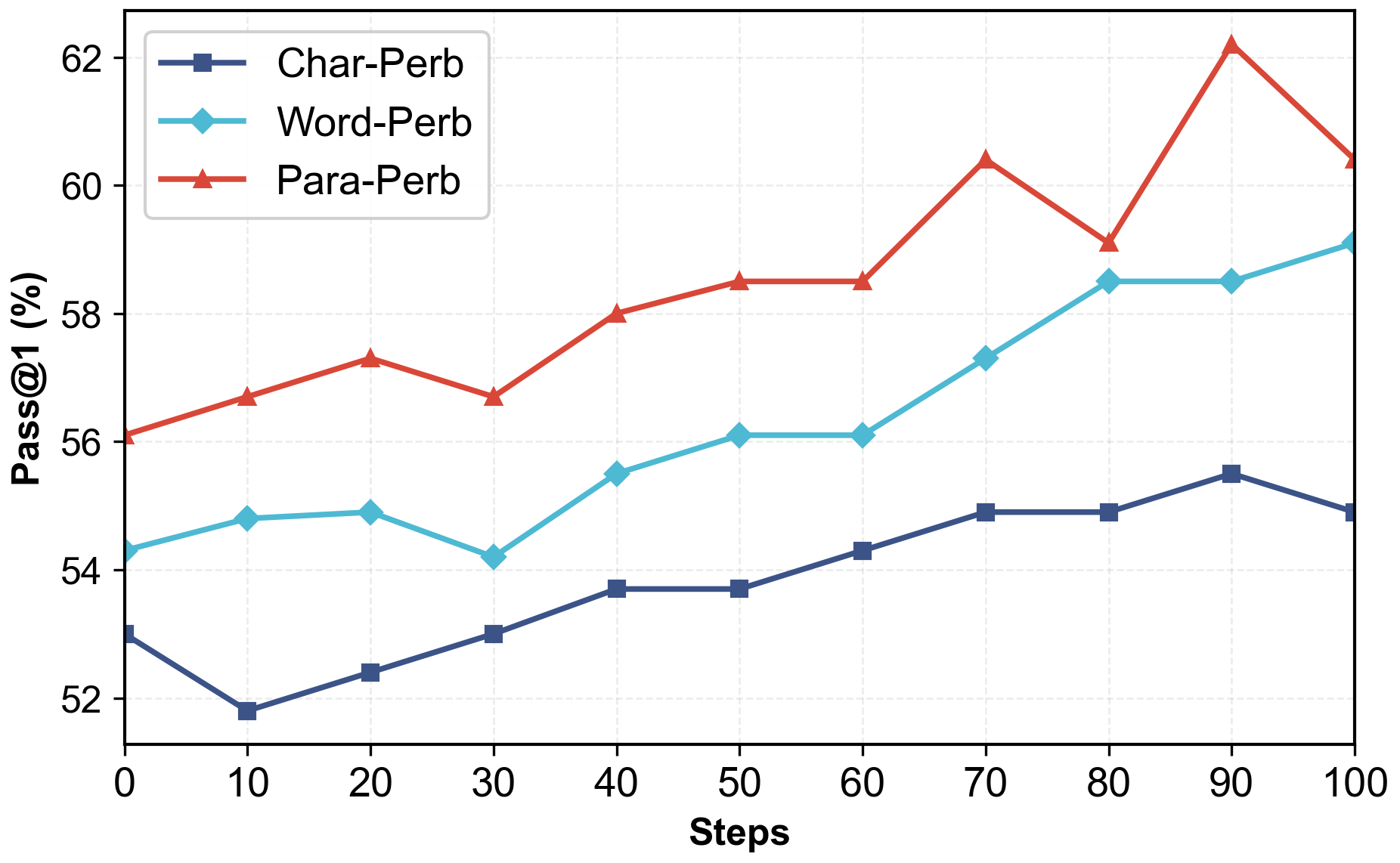} 
  \caption{Evolution of the number of steps for NTRL-Code using DeepSeek-Coder-1.3B on the HumanEval.}
  \label{fig:step_ablation}
\end{figure}

\noindent Figure~\ref{fig:step_ablation} illustrates the performance evolution of NTRL-Code steps ranging from 0 to 100, using the DeepSeek-Coder-1.3B model on the HumanEval dataset with three types of perturbations. The results clearly demonstrate that Pass@1 score steadily increases, indicating continuous improvement and refinement in the model's performance.
Furthermore, NTRL-Code achieves the most significant performance gain under paragraph-level perturbations, while the improvement under character-level perturbations is relatively minor. This suggests that fine-grained noise is more sensitive to LLM generation, introducing greater challenges during the test-time training process.

\subsection{Results on Mixed-Type Perturbations}
\noindent To evaluate whether NTRL-Code is limited to isolated perturbation types, we further test mixed-type perturbations, where multiple perturbation operators are composed within the same prompt. This setting better approximates practical user inputs, in which typos, word substitutions, and paraphrastic variations may co-occur.
Table~\ref{table:mixed_noise} presents the results on the HumanEval dataset using DeepSeek-Coder-1.3B with four types of mixed perturbations. Our findings reveal that the performance of the base model deteriorates as the perturbations become more complex. Nevertheless, NTRL-Code consistently enhances the base model's performance, even achieving the largest improvement (6\%) when all three types of perturbations are added.

\begin{table}[!h]
\centering
\setlength{\tabcolsep}{2.3pt}
\renewcommand{\arraystretch}{1.15} 
\begin{tabular}{>{\centering\arraybackslash}m{3.5cm}|cc}
\toprule
Perturbation Type    & Original   & NTRL-Code   \\
\midrule
Word+Para &$\text{48.1}$ &$\text{53.0}_{+4.9}$  \\
Char+Para &$\text{51.2}$ &$\text{54.3}_{+3.1}$  \\
Char+Word &$\text{47.5}$ &$\text{50.0}_{+2.5}$  \\
Char+Word+Para &$\text{47.0}$ &$\text{53.0}_{+6.0}$  \\
\bottomrule
\end{tabular}
\caption{Performance of the NTRL-Code with mixed-type perturbations.}
\label{table:mixed_noise}
\end{table}

\subsection{Generalization Analysis}
\label{sec:general_analysis}
\noindent To evaluate the generalization ability of our NTRL-Code framework, we test the performance of the trained model on mismatched datasets or perturbations. Specifically, our experiments consist of two parts:  
\begin{enumerate}
    \item \textit{Cross-Dataset Validation}: We fine-tune DeepSeek-Coder-1.3B on the HumanEval dataset using the NTRL-Code framework and test its performance on the LeetCode Bench.  
    \item \textit{Cross-Perturbation Validation}: We train the model on one type of noise and validate its performance on a different type of noise.  
\end{enumerate}
The corresponding experimental results are summarized in Table~\ref{table:general_analysis}. For cross-dataset validation, the results demonstrate that NTRL-Code improves performance across all types of perturbations, indicating strong generalization to previously unseen datasets.  
Additionally, in the cross-perturbation scenario, NTRL-Code still provides improvements, further highlighting its robust generalization ability even when the model encounters mismatched noise types.
These results suggest that NTRL-Code does not merely memorize the perturbation pattern seen during adaptation. Instead, the decoupled objective encourages mappings that transfer across perturbation types and datasets.

\begin{table}[!h]
\centering
\setlength{\tabcolsep}{2.3pt}
\renewcommand{\arraystretch}{1.15} 
\begin{tabular}{>{\centering\arraybackslash}m{3.5cm}|cc}
\toprule
\multicolumn{3}{c}{\textbf{Cross-Dataset (HumanEval $\longrightarrow$ LeetCode)}} \\
\midrule
Perturbation Type    & Original   & NTRL-Code   \\
\midrule
Char-Perb &$\text{3.3}$ &$\text{5.0}_{+1.7}$  \\
Word-Perb &$\text{5.0}$ &$\text{6.1}_{+1.1}$  \\
Para-Perb &$\text{5.0}$ &$\text{6.7}_{+1.7}$  \\
\midrule
\multicolumn{3}{c}{\textbf{Cross-Perturbation (HumanEval)}} \\
\midrule
Para$\rightarrow$Char &$\text{53.0}$ &$\text{53.7}_{+0.7}$  \\
Para$\rightarrow$Word &$\text{54.3}$ &$\text{56.1}_{+1.8}$  \\
\bottomrule
\end{tabular}
\caption{Generalization analysis of the NTRL-Code (the base model is DeepSeek-Coder-1.3B).}
\label{table:general_analysis}
\end{table}

\section{Conclusion}
\noindent In this work, we present NTRL-Code, a label-free test-time reinforcement learning framework for improving code LLM robustness under noisy prompts. NTRL-Code decouples proxy-target estimation from policy optimization by estimating structure-aware targets from rechecked denoised rollouts while updating the model on the original noisy prompts with a hybrid reward. Experiments across three benchmarks, multiple perturbation types, and several base models show consistent robustness gains, while ablations support the roles of self-denoising, AST-based aggregation, and the hybrid reward.

\section*{Limitations}
\noindent The limitations of this work can be summarized as follows.

\noindent \textbf{Noise scope.}
This work focuses on character-level, word-level, paragraph-level, and mixed perturbations. These perturbations cover common surface-form and paraphrastic variations, but they do not exhaust all realistic user-side noise. For example, real coding requests may contain missing input/output specifications, logically inconsistent requirements, multi-task instructions, or ambiguous constraints. Handling such requirement-level noise may require clarification mechanisms or uncertainty-aware refusal, which are beyond the scope of this work.

\noindent \textbf{Proxy-target reliability.}
NTRL-Code estimates proxy targets from model-generated candidates without using executable tests during optimization. Therefore, the selected AST-majority target is not guaranteed to be correct. Failure can occur when the model samples share a common structural bug, when the denoising step changes an important requirement, or when multiple correct programs have different AST structures and are split across clusters. Our design mitigates this through conservative denoising recheck, AST canonicalization, and hybrid rewards, but execution-based verification could further improve reliability.

\noindent \textbf{Computational cost.}
NTRL-Code requires multiple rollouts and GRPO updates during adaptation, making it more expensive than standard single-pass inference. We therefore view it as an asynchronous adaptation method: unlabeled noisy prompts can be batched to update a shadow policy in the background, while the active model serves user requests normally. This setting reduces user-facing latency but still requires additional compute resources.

\noindent \textbf{Closed-source models.}
Our experiments focus on open-source code LLMs because NTRL-Code requires access to model parameters for test-time policy updates. Closed-source APIs can be evaluated for noise sensitivity, but the proposed parameter-update framework cannot be directly applied to them. Developing black-box or training-free variants for closed-source models is an important future direction.

\section*{Acknowledgements}
\noindent The work described in this paper was supported by the Research Grants Council of the Hong Kong Special Administrative Region, China, under Project T45-401/22-N.
AI assistants were used to support the implementation of noise simulation procedures. All experimental designs, analyses, and conclusions were conducted and verified by the authors.

\bibliography{custom}

\clearpage
\appendix

\section{Recap of Test-time Reinforcement Learning}\label{sec:recap_ttrl}
\noindent Given an input prompt $x$, a policy model $\pi_{\theta}(\cdot|x)$, parameterized by $\theta$, is executed multiple times to produce a sequence of candidate outputs $\{y_1, y_2, \cdots, y_N\}$. A proxy ground-truth $y^*$ is then estimated through majority voting, as shown in Eq.~\eqref{eq:majority_voting}:
\begin{equation}
    y^*
= \arg\max_{y\in\{y_1,\ldots,y_N\}}
\sum_{i=1}^{N} \mathbf{1}\{y_i = y\} \, .
\label{eq:majority_voting}
\end{equation}
Afterwards, the current policy model can be updated via the Group Relative Policy Optimization (GRPO)~\cite{shao2024deepseekmath} and the specific objective is given as in Eq.~\eqref{eq:grpo},
\begin{equation}
\begin{aligned}
\mathcal{J}_{\mathrm{GRPO}}(\theta)&
=\mathbb{E}\bigg[\frac{1}{N}\sum_{i=1}^{N}\frac{1}{|y_i|}\sum_{t=1}^{|y_i|}
\min\!\big(r_{i,t}(\theta)\hat A_{i,t},\,\\
&\tilde r_{i,t}(\theta)\hat A_{i,t}\big)
-\beta\,\mathrm{D}_{\mathrm{KL}}\!\left(\pi_{\theta}\,\|\,\pi_{\mathrm{ref}}\right)\bigg],
\end{aligned}
\label{eq:grpo}
\end{equation}
where $r_{i,t}(\theta)$ and $\tilde r_{i,t}(\theta)$ are defined in Eq.~\eqref{eq:r_i_t}, and $\mathrm{D}_\mathrm{KL}$ represents the KL divergence.
\begin{equation}
\begin{aligned}
r_{i,t}(\theta)&=
\frac{\pi_{\theta}(y_{i,t}\mid q,y_{i,<t})}{\pi_{\text{old}}(y_{i,t}\mid q,y_{i,<t})}, \\
\tilde r_{i,t}(\theta)&=\mathrm{clip}\!\big(r_{i,t}(\theta),\,1-\varepsilon,\,1+\varepsilon\big).
\end{aligned}
\label{eq:r_i_t}
\end{equation}
The advantage $\hat A_{i,t}$ measures the relative performance within the group and is computed as shown in Eq.~\eqref{eq:advantage}. $r(\cdot, \cdot)$ refers to the reward function.
\begin{equation}
\begin{aligned}
\hat A_{i,t} = \frac{r(y^*, y_i) - \mathrm{mean}_j(r(y^*,y_j))}{\mathrm{std}_j(r(y^*,y_j))}
\end{aligned}
\label{eq:advantage}
\end{equation}

\section{Implementation Details}

\subsection{Details of Perturbation Simulation} \label{sec:perb_details}
\noindent We construct noisy code generation benchmarks based on three datasets: LeetCode-Bench~\cite{guo2024deepseek}, HumanEval~\cite{chen2021evaluating}, and MBPP~\cite{austin2021program}. Note that the datasets and models used in our experiments are publicly available and are subject to their original licenses. Specifically, LeetCode-Bench consists of 180 problems, categorized into three difficulty levels: Easy, Medium, and Hard. For the HumanEval benchmark, we use its Python subset to simplify the testing pipeline, which includes a total of 164 samples. For the MBPP benchmark, we follow the setup in DeepSeek-Coder~\cite{guo2024deepseek}, selecting task IDs ranging from 11 to 510 for testing.\\
1. \textit{Character-level perturbation.} We simulate common keyboard typos, which often occur when users input long content. This is implemented using the \textit{KeyboardAug} class from the \textbf{nlpaug} library. \\
2. \textit{Word-level perturbation.} We replace certain words in the prompt with their synonyms or near-synonyms. For example, replacing “consecutive” with “sequent” reflects potential errors by non-native speakers who might be unclear about the correct usage of common words or phrases. This is implemented using the \textit{SynonymAug} class from the \textbf{nlpaug} library.\\
3. \textit{Paragraph-level perturbation.} We achieve this by applying rephrasing operations using a powerful language model (e.g. \textit{Qwen3-Coder-480B-A35B-Instruct}). The specific prompt is shown in Figure~\ref{fig:pt_rephrase}.

\subsection{Experiment Details for NTRL-Code}
\noindent All experiments are conducted on a single node equipped with 8$\times$NVIDIA H200 (140\,GB) GPUs.
We adopt a maximum response length of 3072 tokens. The optimizer uses a base learning rate of $5\times10^{-7}$ with cosine decay and warmup; the warmup ratio is set to 0.03 of total steps. During the training stage, rollouts are sampled with temperature 0.6. The mini-batch size and the macro-batch size per GPU are set as 1 and 2, respectively.
For the self-denoising agent, we use deterministic decoding and the denoised prompt has a maximum length of 512 tokens before subsequent rollouts.
For inference, we use greedy decoding with $\tau=0$ for code generation and cap the maximum response length at 1024 tokens.
For AST-parsable programs, we canonicalize identifiers and serialize the AST using Python ast.dump() function (excluding attributes and annotations). We compute an MD5 hash of the serialized canonical tree and use exact hash matching for structural clustering. For unparsable programs, we use token-level Jaccard similarity after code normalization. The lexical clustering threshold is set to \(\epsilon=0.92\).
We compute the length-change ratio between the original prompt and the denoised prompt and set \(\tau_{\mathrm{len}}=0.5\). If the ratio exceeds this threshold, the denoised prompt is rejected and the original prompt is used for target-estimation rollouts.

\begin{table*}[!h]
\centering
\setlength{\tabcolsep}{2.6pt}
\renewcommand{\arraystretch}{1.15} 
\scalebox{0.72}{
\begin{tabular}{>{\centering\arraybackslash}m{3.3cm}cccc|cccc|cccc}
\toprule
\multirow{2}{=}{\centering\textbf{Models}} & \multicolumn{4}{c|}{\textbf{HumanEval}}  & \multicolumn{4}{c|}{\textbf{LeetCode Bench}}  & \multicolumn{4}{c}{\textbf{MBPP}}\\
\cmidrule(lr){2-13}
& \textit{Ori}  & \textit{Char-Perb}    & \textit{Word-Perb}   & \textit{Para-Perb}  & \textit{Ori}   & \textit{Char-Perb}    & \textit{Word-Perb}   & \textit{Para-Perb}  & \textit{Ori} & \textit{Char-Perb}    & \textit{Word-Perb}   & \textit{Para-Perb} \\
\midrule
DS-Coder-1.3B &$\text{64.0}$ &$\text{53.0}_{-11.0}$  &$\text{54.3}_{-9.7}$ &$\text{56.1}_{-7.9}$ &$\text{5.6}$ &$\text{3.3}_{-2.3}$ &$\text{5.0}_{-0.6}$  &$\text{5.0}_{-0.6}$  &$\text{47.4}$ &$\text{44.6}_{-2.8}$ &$\text{43.8}_{-3.6}$ &$\text{46.6}_{-0.8}$  \\
DS-Coder-6.7B &$\text{81.7}$ &$\text{62.8}_{-18.9}$  &$\text{64.0}_{-17.7}$ &$\text{66.5}_{-15.2}$ &$\text{19.4}$ &$\text{16.1}_{-3.3}$ &$\text{15.0}_{-4.4}$  &$\text{16.1}_{-3.3}$  &$\text{64.0}$ &$\text{60.0}_{-4.0}$ &$\text{57.8}_{-6.2}$ &$\text{61.6}_{-2.4}$  \\
Qwen2.5-coder-32B &$\text{91.5}$ &$\text{79.9}_{-11.6}$ &$\text{81.7}_{-9.8}$  &$\text{69.5}_{-22.0}$ &$\text{75.0}$ &$\text{63.9}_{-11.1}$  &$\text{63.3}_{-11.7}$  &$\text{67.2}_{-7.8}$  &$\text{81.2}$ &$\text{80.0}_{-1.2}$ &$\text{79.2}_{-2.0}$ &$\text{78.2}_{-3.0}$  \\
Qwen3-coder-30B &$\text{94.5}$ &$\text{82.3}_{-12.2}$  &$\text{87.2}_{-7.3}$  &$\text{69.5}_{-25.0}$ &$\text{65.6}$ &$\text{64.4}_{-1.2}$  &$\text{63.3}_{-2.3}$  &$\text{62.2}_{-3.4}$  &$\text{78.2}$ &$\text{77.4}_{-0.8}$ &$\text{77.6}_{-0.6}$ &$\text{77.2}_{-1.0}$  \\
Qwen3-Next-80B &$\text{92.7}$ &$\text{80.5}_{-12.2}$ &$\text{83.5}_{-9.2}$  &$\text{69.5}_{-23.2}$  &$\text{60.6}$ &$\text{56.7}_{-3.9}$  &$\text{60.0}_{-0.6}$  &$\text{58.9}_{-1.7}$  &$\text{82.6}$ &$\text{82.0}_{-0.6}$ &$\text{81.0}_{-1.6}$ &$\text{82.2}_{-0.4}$   \\
Qwen3-235B &$\text{95.7}$ &$\text{85.4}_{-10.3}$ &$\text{87.8}_{-7.9}$ &$\text{73.8}_{-21.9}$  &$\text{82.8}$ &$\text{72.8}_{-10.0}$  &$\text{76.1}_{-6.7}$  &$\text{78.3}_{-4.5}$  &$\text{76.2}$ &$\text{61.0}_{-15.2}$ &$\text{69.0}_{-7.2}$ &$\text{71.0}_{-5.2}$  \\
\bottomrule
\end{tabular}
}
\caption{Pass@1 accuracy on the original benchmarks (\textit{Ori}) and three types of perturbed benchmarks (\textit{Char-Perb}, \textit{Word-Perb}, \textit{Para-Perb}) across the HumanEval, LeetCode-Bench, and MBPP datasets. Here, DS-Coder refers to DeepSeek-Coder.}
\label{table:noisy_bench}
\end{table*}

\begin{table}[!h]
\centering
\setlength{\tabcolsep}{2.3pt}
\renewcommand{\arraystretch}{1.15} 
\scalebox{0.83}{
\begin{tabular}{>{\centering\arraybackslash}m{3.0cm}|cccc}
\toprule
\textbf{Model} & \textbf{Clean} & \textbf{Char} & \textbf{Word} & \textbf{Para} \\
\midrule
GPT-4o & $\text{91.5}$ & $\text{85.4}_{-6.1}$ & $\text{84.1}_{-7.4}$ & $\text{90.2}_{-1.3}$ \\
GPT-5 & $\text{97.6}$ & $\text{87.8}_{-9.8}$ & $\text{92.7}_{-4.9}$ & $\text{96.3}_{-1.3}$ \\
GPT-5-Codex & $\text{97.6}$ & $\text{90.2}_{-7.4}$ & $\text{90.2}_{-7.4}$ & $\text{97.6}_{0.0}$ \\
Claude-Opus-4.5 & $\text{98.2}$ & $\text{95.1}_{-3.1}$ & $\text{97.6}_{-0.6}$ & $\text{94.5}_{-3.7}$ \\
\bottomrule
\end{tabular}
}
\caption{Performance of frontier closed-source LLMs on HumanEval under clean and perturbed prompts.}
\label{table:closed_source_noise}
\end{table}

\section{Additional Results}

\subsection{Robustness Evaluation for Code LLMs} \label{sec:prelimiary}
\noindent To conduct a comprehensive evaluation, we select six advanced open-sourced models including DeepSeek-Coder-1.3B, 6.7B~\cite{guo2024deepseek}, Qwen2.5-Coder-32B~\cite{hui2024qwen2}, Qwen3-Coder-30B-A3B~\cite{qwen3technicalreport}, Qwen3-Next-80B-A3B~\cite{qwen3technicalreport}, and Qwen3-235B-A22B~\cite{qwen3technicalreport}. For each model, we evaluate its performance on both the original datasets and the corresponding noisy datasets to observe performance variations under the three types of perturbations. 
Tables~\ref{table:noisy_bench} presents the benchmarking results, where \textit{Char-Perb}, \textit{Word-Perb}, \textit{Para-Perb} represent the character-level, word-level, and paragraph-level perturbations, respectively.
We find that all models experience a noticeable performance drop across the three levels of perturbation, highlighting the necessity of proposing an effective framework to improve their robustness to noisy instructions.

To further validate the influence of prompt perturbations on the closed-source premier LLMs, we also ran experiment on four frontier closed models (GPT-4o, GPT-5, GPT-5-Codex, Claude-Opus-4.5) on HumanEval (Table~\ref{table:closed_source_noise}). We can find that even the strongest frontier models show measurable, non-trivial performance drops under most types of noise (e.g., GPT-5-Codex on char- and word-level noise). Therefore, facial-noise sensitivity is not confined to the lower-tier models but is also present across model tiers.

\subsection{Analysis of the Noise Severity} \label{sec:noise_analysis}
\noindent In this work, we define per-type severity metrics: 
(i) For character noise, the character-level edit distance normalized by prompt length (char-edit ratio).
(ii) For word noise, the fraction of content tokens substituted (word-change fraction).
(iii) For semantic preservation, we embed clean and perturbed prompts with \textit{text-embedding-3-small} and report cosine similarity.
Table~\ref{table:noise_severity} presents computed noise severity metrics across three benchmarks. 
We find that the severity increases monotonically (character < word < paragraph on every benchmark). 
At the same time, the semantic cosine similarity between clean and perturbed prompts stays high, confirming that the perturbations preserve the underlying meaning while varying the surface form. 
Note that the MBPP paragraph char-edit ratio exceeds 1 because MBPP prompts are a single short sentence and the paraphrase substantially expands them, so the normalized edit distance is large by construction, which is consistent with the high 0.826 word-change fraction.

\begin{table}[!h]
\centering
\setlength{\tabcolsep}{2.3pt}
\renewcommand{\arraystretch}{1.15} 
\scalebox{0.74}{
\begin{tabular}{>{\centering\arraybackslash}m{2.5cm}|cccc}
\toprule
\textbf{Benchmark} & \textbf{Noise} & \makecell{\textbf{Char-edit}\\\textbf{Ratio}} & \makecell{\textbf{Word-change}\\\textbf{Fraction}} & \makecell{\textbf{Semantic Cos.}\\\textbf{Mean / Min}} \\
\midrule
\multirow{3}{*}{HumanEval}  & Char & $\text{0.132}$ & $\text{0.353}$ & $\text{0.930 / 0.817}$ \\
 & Word & $\text{0.175}$ & $\text{0.373}$ & $\text{0.960 / 0.880}$ \\
 & Para & $\text{0.425}$ & $\text{0.454}$ & $\text{0.953 / 0.829}$ \\
\midrule
\multirow{3}{*}{LeetCode}  & Char & $\text{0.062}$ & $\text{0.281}$ & $\text{0.912 / 0.831}$ \\
 & Word & $\text{0.119}$ & $\text{0.303}$ & $\text{0.929 / 0.778}$ \\
 & Para & $\text{0.366}$ & $\text{0.456}$ & $\text{0.944 / 0.869}$ \\
\midrule
\multirow{3}{*}{MBPP} & Char & $\text{0.036}$ & $\text{0.095}$ & $\text{0.809 / 0.543}$ \\
 & Word & $\text{0.105}$ & $\text{0.124}$ & $\text{0.865 / 0.632}$ \\
 & Para & $\text{5.524}$ & $\text{0.826}$ & $\text{0.763 / 0.616}$ \\
\bottomrule
\end{tabular}
}
\caption{Computed noise severity metrics across three benchmarks. Character-edit ratio and word-change fraction quantify surface-level perturbation severity, while semantic cosine similarity measures semantic preservation between clean and perturbed prompts.}
\label{table:noise_severity}
\end{table}

\subsection{Effect of Number of Voting Samples} 
\begin{figure}[!h]
  \includegraphics[width=0.48\textwidth]{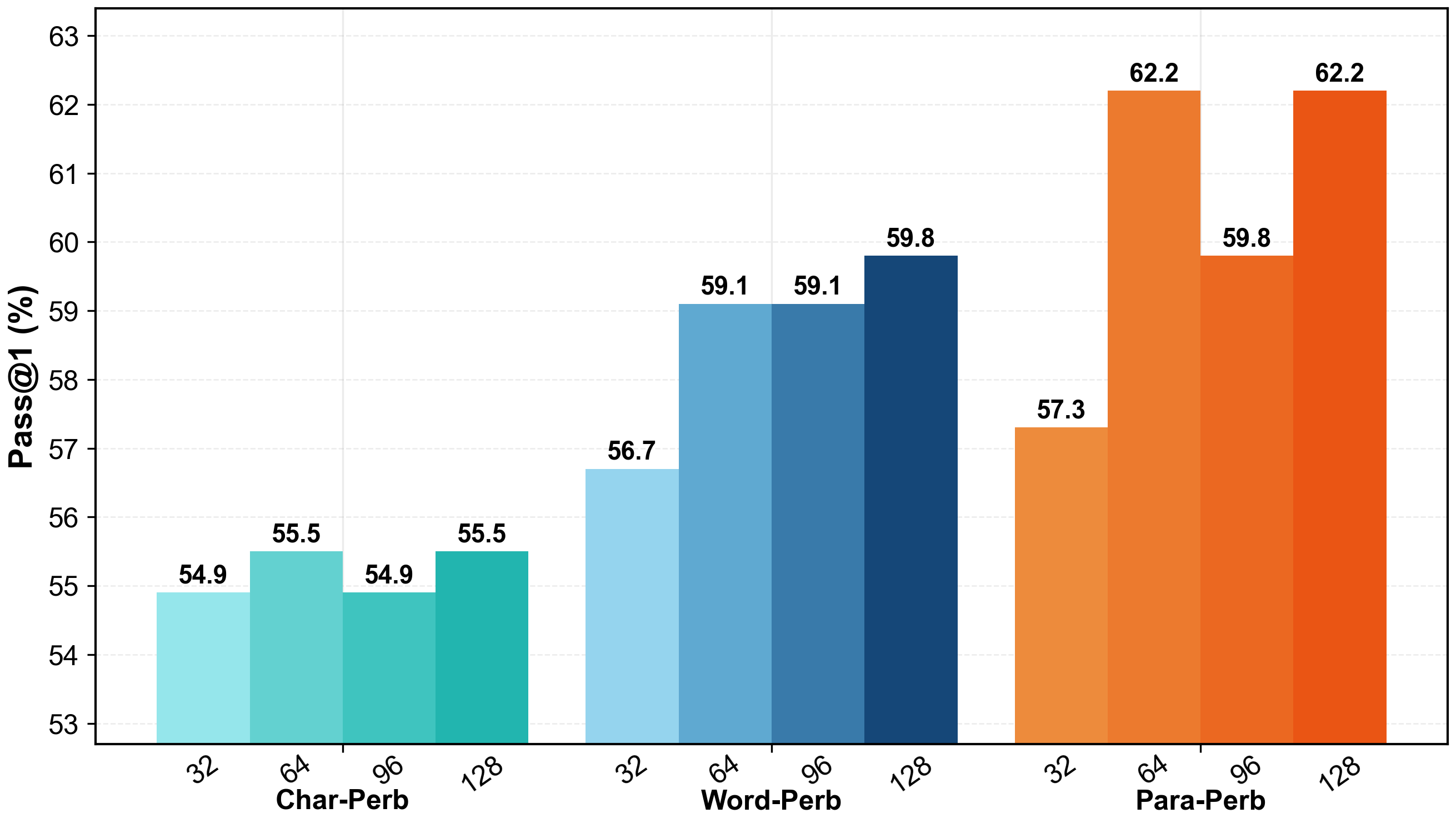} 
  \caption{Performance of NTRL-Code with DeepSeek-Coder-1.3B on HumanEval using 32, 64, 96, and 128 samples for majority voting.}
  \label{fig:ablation_sample}
\end{figure}

\noindent We investigate the impact of the number of samples ($N$) involved in majority voting. Figure~\ref{fig:ablation_sample} shows the performance variation as the sample size increases from 32 to 128 in intervals of 32. Our findings indicate that when $N=32$, the results are generally the worst. While increasing the number of samples provides some benefits, the performance gains tend to diminish as $N$ continues to grow.
Considering the increased computational cost associated with enlarging the majority group, we opt to use 64 samples in most of experiments.

\subsection{Performance on the Clean Dataset}
\noindent Here, we examine the influence of NTRL-Code onto three clean benchmarks using the DeepSeek-Coder-1.3B model. As shown in Table~\ref{table:ntrl_impro}, even without perturbations, NTRL-Code delivers consistent gains with about $3.2\%$ on average across datasets. We attribute these improvements to test-time training effects, which are in accordance with results reported in prior works~\cite{zuo2026ttrl,singh2026ttrv}.

\begin{table}[!h]
\centering
\setlength{\tabcolsep}{2.3pt}
\renewcommand{\arraystretch}{1.15} 
\scalebox{0.83}{
\begin{tabular}{>{\centering\arraybackslash}m{2.5cm}|ccc}
\toprule
& \textbf{HumanEval}    & \textbf{LeetCode Bench}   & \textbf{MBPP}   \\
\midrule
Original &$\text{64.0}$ &$\text{5.6}$ &$\text{47.4}$  \\
NTRL-Code &$\text{67.1}_{+3.1}$ &$\text{8.3}_{+2.7}$ &$\text{51.2}_{+3.8}$ \\
\bottomrule
\end{tabular}
}
\caption{Performance improvements of NTRL-Code on three clean benchmarks using the DeepSeek-Coder-1.3B model.}
\label{table:ntrl_impro}
\end{table}

\subsection{Hyperparameter Sensitivity Analysis}
\noindent To verify the choice of $\tau_{\text{len}}$, we swept $\tau_{\text{len}} \in \{0.3, 0.4, 0.5, 0.6, 0.7\}$ on HumanEval character-noise (DeepSeek-Coder-1.3B) and report pass@1 together with the revert rate (fraction of denoised prompts rejected by the guardrail) as shown in Table~\ref{table:tau_len_sensitivity}.

\begin{table}[!h]
\centering
\setlength{\tabcolsep}{2.3pt}
\renewcommand{\arraystretch}{1.15} 
\scalebox{0.83}{
\begin{tabular}{>{\centering\arraybackslash}m{2.5cm}|ccccc}
\toprule
$\tau_{\text{len}}$ & $\text{0.3}$ & $\text{0.4}$ & $\text{0.5}$ & $\text{0.6}$ & $\text{0.7}$ \\
\midrule
Pass@1 (\%) & $\text{54.3}$ & $\text{55.5}$ & $\text{55.5}$ & $\text{51.8}$ & $\text{51.8}$ \\
Revert rate (\%) & $\text{76.8}$ & $\text{74.4}$ & $\text{73.8}$ & $\text{70.1}$ & $\text{67.7}$ \\
\bottomrule
\end{tabular}
}
\caption{Sensitivity analysis of the length-threshold parameter $\tau_{\text{len}}$ on HumanEval character-noise using DeepSeek-Coder-1.3B.}
\label{table:tau_len_sensitivity}
\end{table}

To observe the stability for the lexical clustering threshold ($\epsilon$) on HumanEval (DeepSeek-Coder-1.3B). We sweep $\epsilon \in \{0.80, 0.85, 0.90, 0.92, 0.95\}$ (Table~\ref{table:epsilon_sensitivity}) and find that pass@1 stays within a 0.6-point band (54.9–55.5\%).

\begin{table}[!h]
\centering
\setlength{\tabcolsep}{2.3pt}
\renewcommand{\arraystretch}{1.15} 
\scalebox{0.83}{
\begin{tabular}{>{\centering\arraybackslash}m{2.5cm}|ccccc}
\toprule
$\epsilon$ & $\text{0.80}$ & $\text{0.85}$ & $\text{0.90}$ & $\text{0.92}$ & $\text{0.95}$ \\
\midrule
Pass@1 (\%) & $\text{54.9}$ & $\text{54.9}$ & $\text{55.2}$ & $\text{55.5}$ & $\text{55.5}$ \\
\bottomrule
\end{tabular}
}
\caption{Sensitivity analysis of the lexical clustering threshold $\epsilon$ on HumanEval using DeepSeek-Coder-1.3B.}
\label{table:epsilon_sensitivity}
\end{table}

We also conduct an experiment on NTRL-Code (DeepSeek-Coder-1.3B, HumanEval character-noise) under four reward-weight settings spanning the full range from equal weighting to similarity-only, and evaluate each trained model's pass@1 on character-level noise test sets. Experimental results on Table~\ref{table:reward_weight_sensitivity} show that the default $(\alpha,\beta)=(0.5,0.25)$ gives the best performance, indicating the proposed hybrid reward benefits from balancing similarity-based guidance with the additional reward components, rather than relying exclusively on similarity or assigning equal weights to all terms.

\begin{table}[!h]
\centering
\setlength{\tabcolsep}{2.3pt}
\renewcommand{\arraystretch}{1.15} 
\scalebox{0.83}{
\begin{tabular}{>{\centering\arraybackslash}m{3.2cm}|cc}
\toprule
\textbf{Setting} & $(\alpha,\beta)$ & \makecell{\textbf{Character}\\\textbf{Pass@1}} \\
\midrule
default & $\text{(0.50, 0.25)}$ & $\textbf{\text{55.5}}$ \\
equal weights & $\text{(0.34, 0.33)}$ & $\text{51.8}$ \\
similarity-heavy & $\text{(0.70, 0.15)}$ & $\text{52.4}$ \\
similarity-only & $\text{(1.00, 0.00)}$ & $\text{50.6}$ \\
\bottomrule
\end{tabular}
}
\caption{Sensitivity analysis of reward-weight settings on NTRL-Code with DeepSeek-Coder-1.3B under HumanEval character-level noise.}
\label{table:reward_weight_sensitivity}
\end{table}

We further investigated the equal averaging of the normalized Levenshtein distance and Jaccard similarity in Eq.~\ref{eq:sim_overall}. Specifically, we swept the Levenshtein weight $w_{\text{lev}} \in \{0.4, 0.5, 0.8\}$, where $w_{\text{lev}} = 0.5$ corresponds to the equal-weight setting. The results confirm that the equal-weight configuration performs best among the tested alternatives.

\begin{table}[!h]
\centering
\setlength{\tabcolsep}{2.3pt}
\renewcommand{\arraystretch}{1.15} 
\scalebox{0.83}{
\begin{tabular}{>{\centering\arraybackslash}m{4.8cm}|ccc}
\toprule
\makecell{\textbf{Levenshtein Weight} $w_{\text{lev}}$} & $\text{0.4}$ & $\text{0.5}$ & $\text{0.8}$ \\
\midrule
pass@1 under Char-Perb & $\text{54.9}$ & $\mathbf{55.5}$ & $\text{54.3}$ \\
\bottomrule
\end{tabular}
}
\caption{Sensitivity analysis of the Levenshtein weight $w_{\text{lev}}$ in Eq.~\ref{eq:sim_overall}.}
\label{table:levenshtein_weight_sensitivity}
\end{table}

Finally, we studied the sensitivity to the step/window ratio in Eq.~\ref{eq:slide_w_s} by testing window/step ratios of 0.25, 0.5, and 1.0. The default ratio of 0.5 again achieves the best pass@1.

\begin{table}[!h]
\centering
\setlength{\tabcolsep}{2.3pt}
\renewcommand{\arraystretch}{1.15} 
\scalebox{0.83}{
\begin{tabular}{>{\centering\arraybackslash}m{4.2cm}|ccc}
\toprule
\makecell{\textbf{Step/Window Ratio}\\\textbf{(Eq. 6)}} & $\text{0.25}$ & $\text{0.5}$ & $\text{1.0}$ \\
\midrule
pass@1 under Char-Perb & $\text{54.3}$ & $\mathbf{55.5}$ & $\text{54.3}$ \\
\bottomrule
\end{tabular}
}
\caption{Sensitivity analysis of the step/window ratio in Eq.~\ref{eq:slide_w_s}.}
\label{table:step_window_ratio_sensitivity}
\end{table}

\subsection{Additional Ablation Results}
\noindent To further validate the effectiveness of self-denoising module, we repeated the self-denoising ablation on the larger DeepSeek-Coder-6.7B in Table~\ref{table:self_denoising_ablation_67b}. Specifically, we evaluate on HumanEval with word-level perturbation and compare the base model against NTRL-Code with and without the self-denoising (SD) branch. The trend observed at 1.3B still holds at 6.7B. NTRL-Code improves over the base model and adding the self-denoising branch gives a further consistent gain.

\begin{table}[!h]
\centering
\setlength{\tabcolsep}{2.3pt}
\renewcommand{\arraystretch}{1.15} 
\scalebox{0.83}{
\begin{tabular}{>{\centering\arraybackslash}m{4.0cm}|c}
\toprule
\textbf{HumanEval, Word-Perb} & \textbf{Pass@1} \\
\midrule
Original (base 6.7B) & $\text{64.0}$ \\
NTRL-Code w/o SD & $\text{66.5}$ \\
NTRL-Code w/ SD & $\text{67.1}$ \\
\bottomrule
\end{tabular}
}
\caption{Self-denoising ablation on DeepSeek-Coder-6.7B under HumanEval word-level perturbation.}
\label{table:self_denoising_ablation_67b}
\end{table}

\subsection{Proxy Target Analysis}
\noindent Here, we analyze the correlation between the proxy target and execution-based correctness. Specifically, we run the AST-canonicalization majority vote to obtain the proxy target, and then execute the candidates and the proxy target against the benchmark test suites.

\noindent \textbf{Execution audit.} For each HumanEval problem, we sampled 32 candidates (DeepSeek-Coder-1.3B), constructed the proxy target via structural aggregation, and executed proxy and candidates against the benchmark tests. We report below metrics and the corresponding results is shown in Table~\ref{table:execution_audit_ast_proxy}.
\begin{itemize}
    \item \textit{proxy pass rate:} Proxy target passes all tests; 
    \item \textit{candidate pass rate:} Expected pass rate of a uniformly random candidate — the gap to proxy pass rate is the selection gain from aggregation; 
    \item \textit{reward–correctness corr:} Per-candidate Pearson correlation between proxy reward and passing; 
    \item \textit{consensus–correctness corr:} Per-problem correlation between majority ratio and proxy correctness, probing the "consistent-but-incorrect" risk; 
\end{itemize}

\begin{table}[!h]
\centering
\setlength{\tabcolsep}{2.3pt}
\renewcommand{\arraystretch}{1.15} 
\scalebox{0.75}{
\begin{tabular}{>{\centering\arraybackslash}m{2.3cm}|cccc}
\toprule
\textbf{Condition} 
& \makecell{\textbf{Proxy}\\\textbf{Pass Rate}} 
& \makecell{\textbf{Candidate}\\\textbf{Pass Rate}} 
& \makecell{\textbf{Reward--}\\\textbf{Correctness}\\\textbf{Corr.}} 
& \makecell{\textbf{Consensus--}\\\textbf{Correctness}\\\textbf{Corr.}} \\
\midrule
clean & $\text{65.2\%}$ & $\text{55.1\%}$ & $\text{0.37}$ & $\text{0.37}$ \\
character & $\text{54.3\%}$ & $\text{44.9\%}$ & $\text{0.38}$ & $\text{0.36}$ \\
word & $\text{56.7\%}$ & $\text{44.2\%}$ & $\text{0.41}$ & $\text{0.31}$ \\
paragraph & $\text{62.8\%}$ & $\text{55.8\%}$ & $\text{0.41}$ & $\text{0.41}$ \\
\bottomrule
\end{tabular}
}
\caption{Execution audit of the AST-canonicalization majority-vote proxy target on HumanEval using DeepSeek-Coder-1.3B.}
\label{table:execution_audit_ast_proxy}
\end{table}

\subsection{NTRL-Code under Requirement-level Noise}
\noindent To verify the performance of NTRL-Code under requirement-level noise, following the HumanEvalComm~\cite{wu2025humanevalcomm} taxonomy, we created 57 prompts across four categories: incomplete specifications (17), inconsistent requirements (12), ambiguous constraints (15), and multi-task instructions (13). Each prompt includes a clean-intent reference and an annotation of the defect. Because functional correctness alone does not capture whether a model detects ambiguity, states assumptions, or addresses all requested tasks, we use a rubric-based GPT-5 judge to score intent correctness, noise handling, and robustness on a 0–5 scale.

The corresponding results in Table~\ref{table:requirement_level_noise} indicate positive transfer to requirement-level noise, particularly in intent recovery and robustness. However, explicit handling of contradictions and missing information remains limited. We therefore interpret this result as a positive trend rather than conclusive evidence. 

\begin{table}[!h]
\centering
\setlength{\tabcolsep}{2.3pt}
\renewcommand{\arraystretch}{1.15}
\scalebox{0.75}{
\begin{tabular}{>{\centering\arraybackslash}m{4.2cm}|ccc}
\toprule
\makecell{\textbf{Metric}\\\textbf{(mean, 0--5 unless \%)}} 
& \textbf{Base} 
& \textbf{NTRL-Code} 
& $\Delta$ \\
\midrule
intent-correctness & $\text{2.30}$ & $\text{2.67}$ & $\mathbf{+0.37}$ \\
noise-handling & $\text{1.49}$ & $\text{1.61}$ & $\mathbf{+0.12}$ \\
robustness & $\text{3.91}$ & $\text{4.21}$ & $\mathbf{+0.30}$ \\
overall pass rate (\%) & $\text{10.5\% (6/57)}$ & $\text{14.0\% (8/57)}$ & $\mathbf{+3.5}$ \\
\bottomrule
\end{tabular}
}
\caption{Evaluation on a manually constructed 57-prompt requirement-level noise benchmark. Metrics are judged by a rubric-based GPT-5 evaluator.}
\label{table:requirement_level_noise}
\end{table}

\subsection{Cross-language Evaluation}
\noindent We ran the experiment (DeepSeek-Coder-1.3B) on the C++ and JavaScript subsets of HumanEval, under both clean and character-noise conditions. Execution uses g++ / node. Importantly, for C++/JS the Python \texttt{ast} parser fails, so proxy-target aggregation falls back entirely to the normalized-Jaccard clustering path, which also serves as a direct stress test of whether the method still works without a language-specific AST.
Based on the experimental results (Table~\ref{table:cross_language_humaneval}), we can observe that NTRL-Code improves over the base model in all four settings (+1.2 to +5.6 points), which confirms that our method generalizes beyond Python and remains effective even when the AST path is unavailable.
\begin{table}[!h]
\centering
\setlength{\tabcolsep}{2.3pt}
\renewcommand{\arraystretch}{1.15}
\scalebox{0.80}{
\begin{tabular}{>{\centering\arraybackslash}m{2.2cm}|>{\centering\arraybackslash}m{2.4cm}|ccc}
\toprule
\textbf{Language} 
& \textbf{Condition} 
& \makecell{\textbf{Base}\\\textbf{Pass@1}} 
& \makecell{\textbf{NTRL-Code}\\\textbf{Pass@1}} 
& $\Delta$ \\
\midrule
\multirow{2}{*}{C++} 
& clean & $\text{44.7}$ & $\text{46.6}$ & $\text{+1.9}$ \\
& char-perb & $\text{40.4}$ & $\text{46.0}$ & $\text{+5.6}$ \\
\midrule
\multirow{2}{*}{JavaScript} 
& clean & $\text{55.9}$ & $\text{60.9}$ & $\text{+5.0}$ \\
& char-perb & $\text{52.8}$ & $\text{54.0}$ & $\text{+1.2}$ \\
\bottomrule
\end{tabular}
}
\caption{Cross-language evaluation on the C++ and JavaScript subsets of HumanEval using DeepSeek-Coder-1.3B under clean and character-noise conditions.}
\label{table:cross_language_humaneval}
\end{table}

\section{Summary of Prompts} \label{sec:prompt_summary}
\begin{figure}[!h]
  \includegraphics[width=0.48\textwidth]{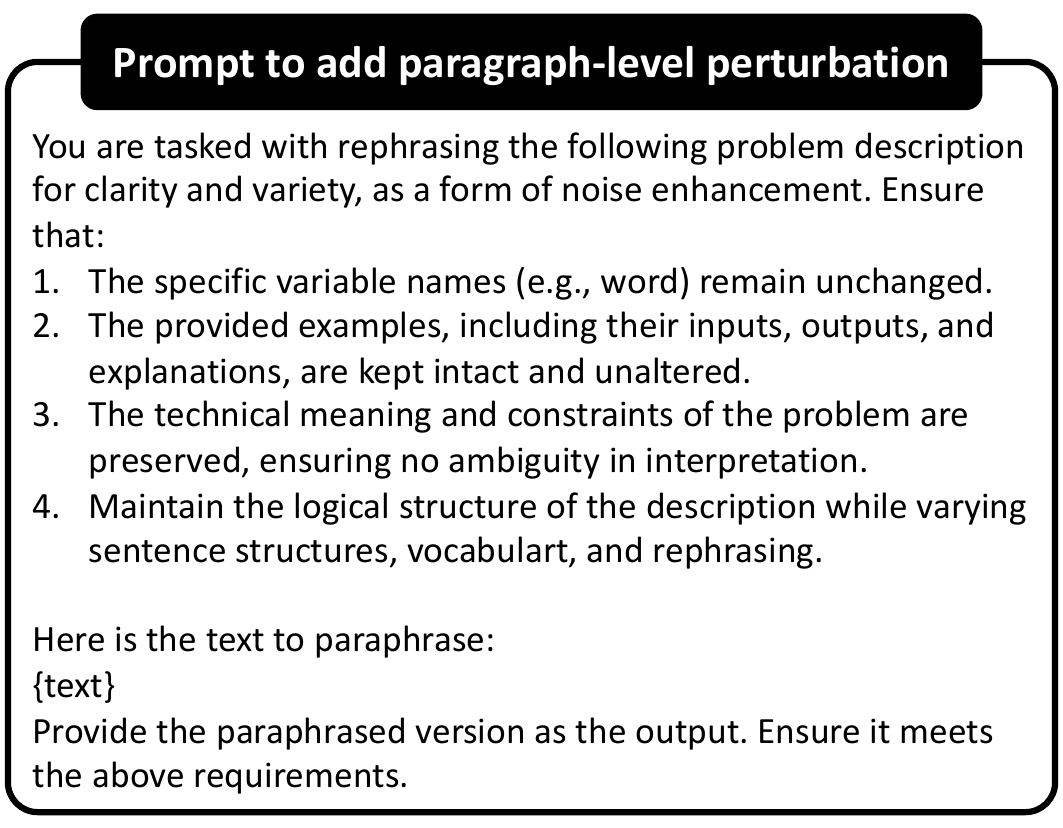} 
  \caption{Prompt used by the Qwen3 agent to perform paragraph-level perturbation through rephrasing.}
  \label{fig:pt_rephrase}
\end{figure}

\begin{figure}[!h]
  \includegraphics[width=0.48\textwidth]{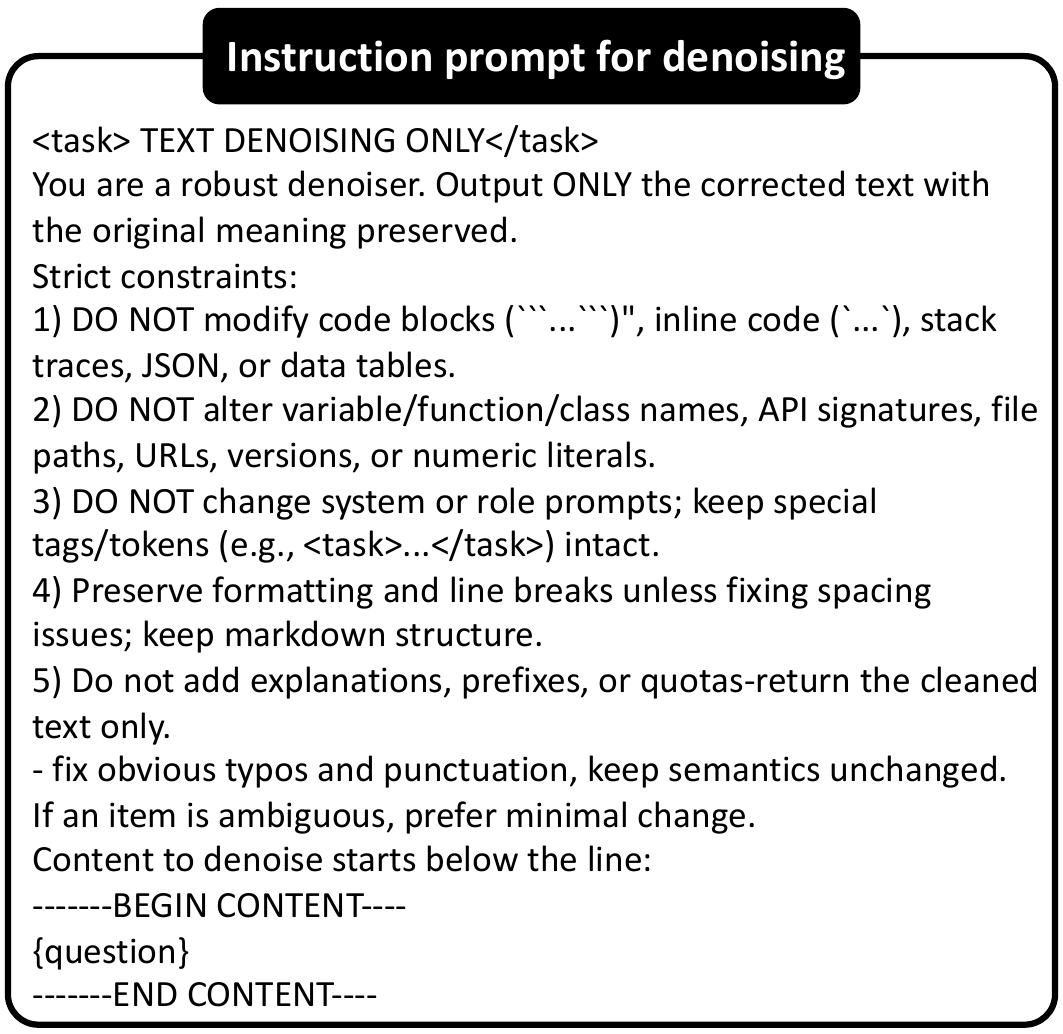} 
  \caption{Prompt to denoise input instruction.}
  \label{fig:pt_SD_ntrl}
\end{figure}

\begin{figure}[!h]
  \includegraphics[width=0.48\textwidth]{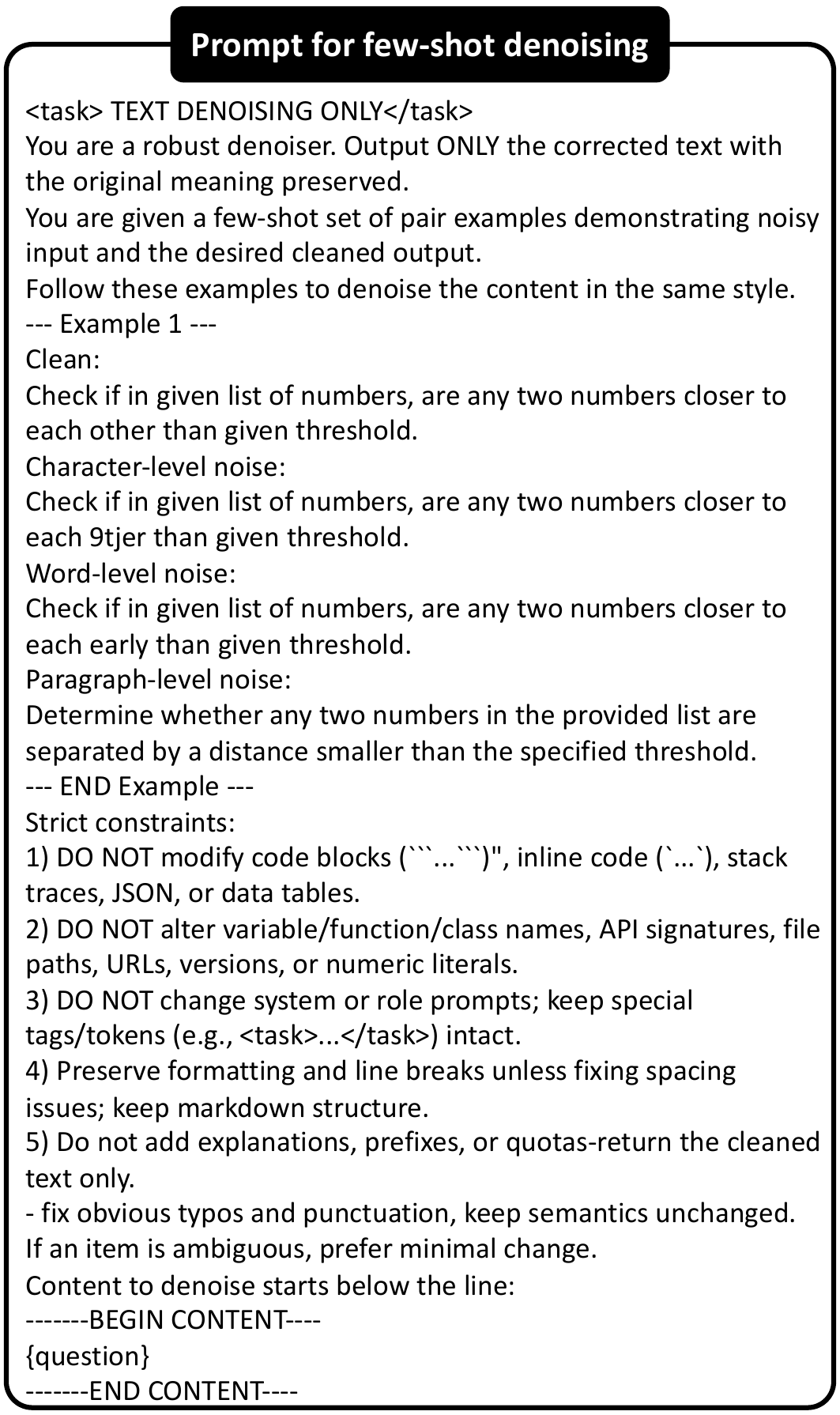} 
  \caption{Prompt to perform few-shot denoising.}
  \label{fig:pt_few_shot}
\end{figure}

\noindent Figures~\ref{fig:pt_rephrase}, \ref{fig:pt_SD_ntrl}, and \ref{fig:pt_few_shot} provide the prompts for adding paragraph-level perturbations, performing self-denoising in NTRL-Code, and conducting few-shot denoising in the baseline method, respectively.


\section{Case Study}
\begin{figure*}
  \includegraphics[width=\textwidth]{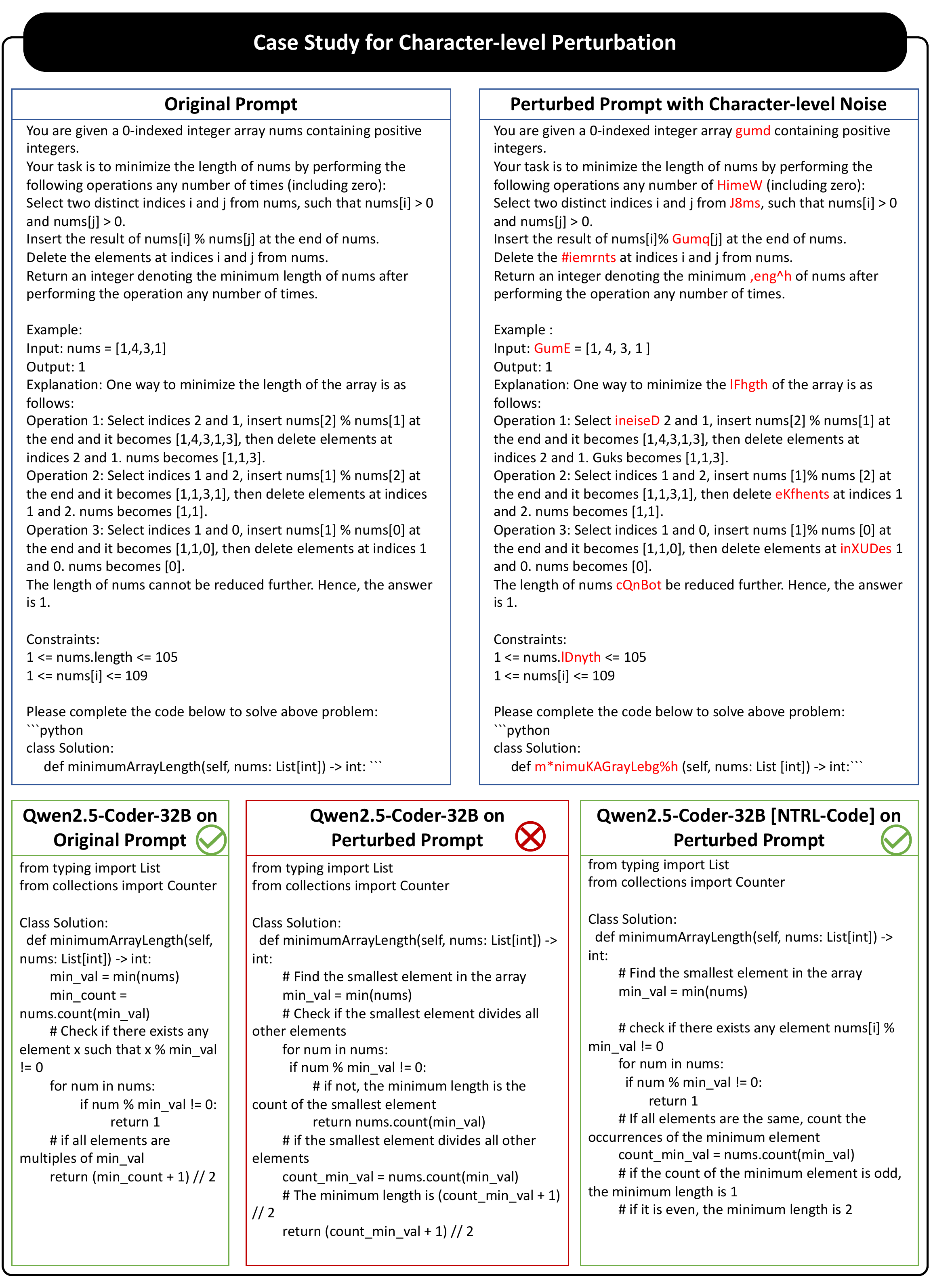} 
  \caption{Case study on character-level perturbation, showcasing the inference results of Qwen2.5-Coder-32B on the original prompt, the perturbed prompt, and Qwen2.5-Coder-32B with NTRL-Code enhancement applied to the perturbed prompt.}
  \label{fig:case1}
\end{figure*}

\begin{figure*}
  \includegraphics[width=\textwidth]{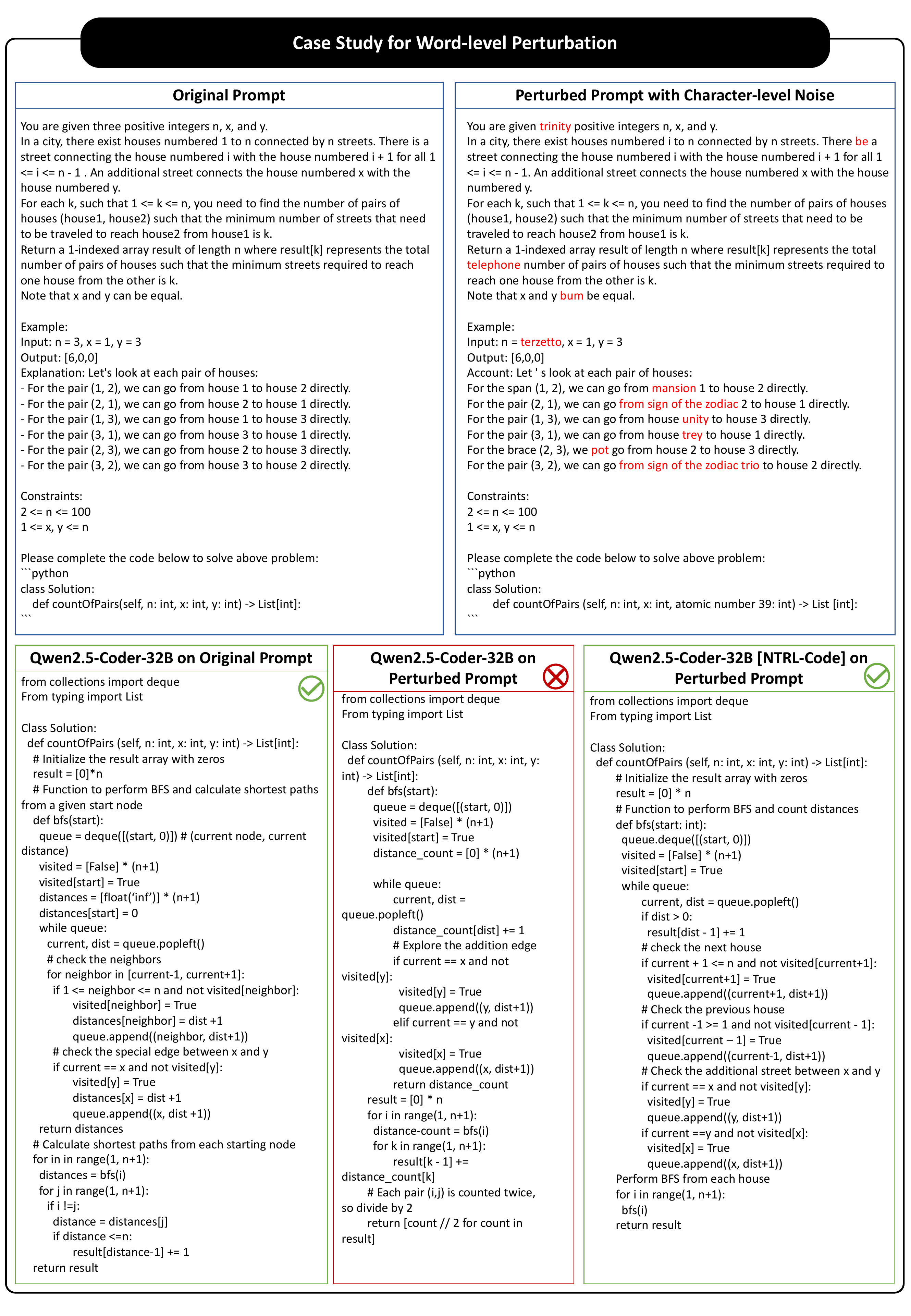} 
  \caption{Case study on word-level perturbation, showcasing the inference results of Qwen2.5-Coder-32B on the original prompt, the perturbed prompt, and Qwen2.5-Coder-32B with NTRL-Code enhancement applied to the perturbed prompt.}
  \label{fig:case2}
\end{figure*}

\begin{figure*}
  \includegraphics[width=\textwidth]{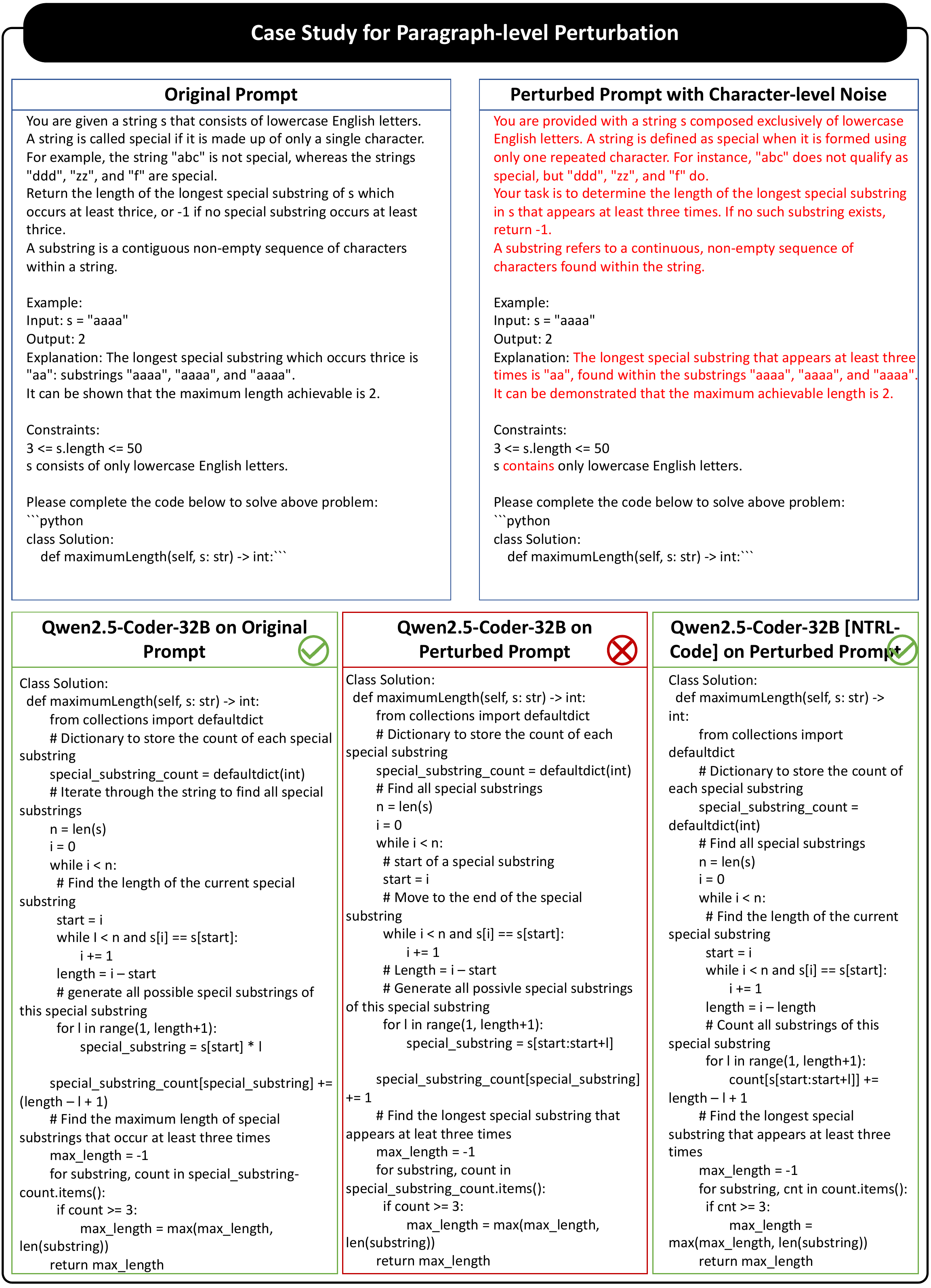} 
  \caption{Case study on paragraph-level perturbation, showcasing the inference results of Qwen2.5-Coder-32B on the original prompt, the perturbed prompt, and Qwen2.5-Coder-32B with NTRL-Code enhancement applied to the perturbed prompt.}
  \label{fig:case3}
\end{figure*}

\noindent Figure~\ref{fig:case1}, \ref{fig:case2}, and \ref{fig:case3} illustrate three cases corresponding to character-level, word-level, and paragraph-level perturbations, respectively. The inference results of Qwen2.5-Coder-32B on the original prompt, the perturbed prompt, and Qwen2.5-Coder-32B with NTRL-Code enhancement applied to the perturbed prompt are presented accordingly.

\section{Discussion}
\subsection{Deployment and Computational Cost}
\noindent NTRL-Code is designed for asynchronous test-time adaptation rather than synchronous per-query optimization. In latency-sensitive applications, the active model can answer user requests using standard inference, while accumulated unlabeled noisy prompts are used to update a shadow policy in the background. After validation, the updated policy can be periodically deployed. Under this setting, the computational cost of denoised rollouts and GRPO updates is amortized over future requests and does not directly increase the response latency of each individual query.

The main computational overhead comes from sampling multiple rollouts. With \(T\) update steps, batch size \(B\), \(K_d\) denoised rollouts for proxy-target estimation, and \(K_n\) noisy rollouts for advantage computation, the rollout cost scales as \(O(TB(K_d+K_n))\). This is more expensive than single-pass inference and may not be suitable when every query requires immediate personalized adaptation. However, compared with iterative self-denoising methods that add synchronous inference-time overhead to each query, NTRL-Code shifts adaptation compute to the background and improves the model parameters persistently.

\subsection{Reliability and Failure Modes of Proxy Targets}
\noindent The AST-based proxy target is an approximate estimate rather than an executable correctness oracle. It can fail when most sampled programs share the same structural bug, when denoising removes or changes an important constraint, or when several correct implementations are structurally diverse and therefore split into different clusters. In these cases, the largest cluster may reflect the model's prior bias instead of the true solution.

NTRL-Code mitigates this risk through three mechanisms. First, the proxy target is estimated from the rechecked denoised prompt, reducing perturbation-induced ambiguity. Second, AST canonicalization groups superficial variants and reduces sensitivity to naming or formatting differences. Third, the hybrid reward avoids relying on a single brittle signal by combining similarity, format validity, and anti-repetition penalties. Nevertheless, without executable feedback, NTRL-Code cannot guarantee that the selected proxy target is always correct. Incorporating lightweight execution-based verification or uncertainty-aware target rejection is a promising direction for future work.

\subsection{Discussion on the Surface-level Metrics in Reward Function}
\noindent In NTRL-Code, two surface-level metrics: Levenshtein and Jaccard are included in the reward computation. We chose these metrics because they are lightweight and require no external encoder. However, these surface metrics are used only as dense reward-shaping signals after target selection, and Jaccard additionally serves as a fallback for unparsable outputs. They do not determine the proxy target. (For parsable programs, the target is selected through canonicalized AST clustering, which is invariant to identifier renaming and formatting differences.)
Moreover, embedding similarity or semantic similarity is not uniformly superior for our purpose: SAGE~\cite{goel2025sage} reports that Jaccard and Levenshtein perform strongly on information-sensitivity and transformation-robustness evaluations, respectively, while prior work~\cite{naik2024limitations} finds that embedding-based similarity metric has only a weak correlation with execution correctness and is therefore not a reliable proxy for functional correctness.

\subsection{Discussion on the AST-based Structural Aggregation}
\noindent In this work, we use MD5 hashes of the AST for structural comparison. We would like to clarify that in our framework, the LLM never sees or reasons about the MD5 hash. The hash is used purely internally as an exact-match key for clustering canonicalized ASTs. We canonicalize the tree (renaming variables/functions to placeholders), serialize it with \textit{ast.dump()} excluding attributes/annotations, and hash the resulting string. Two code snippets are placed in the same cluster iff their canonicalized ASTs are identical. To sum up, the hash is only a cheap, deterministic equality test over normalized structures. It is not a learned representation and is not consumed by the policy. This makes the aggregation step exact and dependency-free (no trained model, no thresholds for the parsable case).
Regarding other structural models like GNNs, these are a reasonable alternative for soft structural similarity. However, one of the trade-offs is that a GNN introduces a trained component, hyperparameters, and non-determinism into a test-time loop. In contrast, exact canonical hashing already handles the dominant failure mode of naive majority voting while keeping the loop parameter-free.

\subsection{Potential Risks}
\noindent While NTRL-Code primarily focuses on enhancing the robustness of Code LLMs, adversarial attacks still pose a potential risk in our approach. For instance, some users may intentionally inject harmful instructions into the model. Without proper security mechanisms in place, such malicious content could be incorporated into the learning process of NTRL-Code, leading the model to be updated based on harmful on-the-fly inputs.


\end{document}